\documentclass{article}%
\usepackage{amsmath}
\usepackage{amsfonts}
\usepackage{amssymb}
\usepackage{graphicx}
\usepackage{algorithmic}
\usepackage{algorithm}%
\providecommand{\U}[1]{\protect\rule{.1in}{.1in}}

\begin{document}

\title{SurvLIME: A method for explaining machine learning survival models}
\author{Maxim S. Kovalev, Lev V. Utkin and Ernest M. Kasimov\\Peter the Great St.Petersburg Polytechnic University (SPbPU)\\St.Petersburg, Russia\\e-mail: lev.utkin@gmail.com, maxkovalev03@gmail.com, kasimov.ernest@gmail.com}
\date{}
\maketitle

\begin{abstract}
A new method called SurvLIME for explaining machine learning survival models
is proposed. It can be viewed as an extension or modification of the
well-known method LIME. The main idea behind the proposed method is to apply
the Cox proportional hazards model to approximate the survival model at the
local area around a test example. The Cox model is used because it considers a
linear combination of the example covariates such that coefficients of the
covariates can be regarded as quantitative impacts on the prediction. Another
idea is to approximate cumulative hazard functions of the explained model and
the Cox model by using a set of perturbed points in a local area around the
point of interest. The method is reduced to solving an unconstrained convex
optimization problem. A lot of numerical experiments demonstrate the SurvLIME efficiency.


\end{abstract}

\section{Introduction}

Many complex problems in various applications are solved by means of deep
machine learning models, in particular deep neural networks, at the present
time. One of the demonstrative examples is the disease diagnosis by the models
on the basis of medical images or another medical information. At the same
time, deep learning models often work as black-box models such that details of
their functioning are often completely unknown. It is difficult to explain in
this case how a certain result or decision is achieved. As a result, the
machine learning models meet some difficulties in their incorporating into
many important applications, for example, into medicine, where doctors have to
have an explanation of a stated diagnosis in order to choose a corresponding
treatment. The lack of the explanation elements in many machine learning
models has motivated development of many methods which could interpret or
explain the deep machine learning algorithm predictions and understand the
decision-making process or the key factors involved in the decision
\cite{Arya-etal-2019,Guidotti-2019,Molnar-2019,Murdoch-etal-2019}.

The methods explaining the black-box machine learning models can be divided
into two main groups: local methods which derive explanation locally around a
test example; global methods which try to explain the overall behavior of the
model. A key component of explanations for models is the contribution of
individual input features. It is assumed that a prediction is explained when
every feature is assigned by some number quantified its impact on the
prediction. One of the first local explanation methods is the Local
Interpretable Model-agnostic Explanations (LIME) \cite{Ribeiro-etal-2016},
which uses simple and easily understandable linear models to locally
approximate the predictions of black-box models. The main intuition of the
LIME is that the explanation may be derived locally from a set of synthetic
examples generated randomly in the neighborhood of the example to be explained
such that every synthetic example has a weight according to its proximity to
the explained example. Moreover, the method uses simple and easily
understandable models like decision rules or linear models to locally or
globally approximate the predictions of black-box models. The method is
agnostic to the black-box model. This means that any details of the black-box
model are unknown. Only its input and the corresponding output are used for
training the explanation model. It is important to mention a work
\cite{Garreau-Luxburg-2020}, which provides a thorough theoretical analysis of
the LIME.

The LIME as well as other methods have been successfully applied to many
machine learning models for explanation. However, to the best of our
knowledge, there is a large class of models for which there are no explanation
methods. These models are applied to problems which take into account survival
aspects of applications. Survival analysis as a basis for these models can be
regarded as a fundamental tool which is used in many applied areas especially
in medicine. The survival models can be divided into three parts: parametric,
nonparametric and semiparametric
\cite{Hosmer-Lemeshow-May-2008,Wang-Li-Reddy-2017}. Most machine learning
models are based on nonparametric and semiparametric survival models. One of
the popular regression model for the analysis of survival data is the
well-known Cox proportional hazards model, which is a semi-parametric model
that calculates effects of observed covariates on the risk of an event
occurring, for example, the death or failure \cite{Cox-1972}. The proportional
hazards assumption in the Cox model means that different examples (patients)
have hazard functions that are proportional, i.e., the ratio of the hazard
functions for two examples with different prognostic factors or covariates is
a constant and does not vary with time. The model assumes that a patient's
log-risk of failure is a linear combination of the example covariates. This is
a very important peculiarity of the Cox model, which will be used below in
explanation models.

A lot of machine learning implementations of survival analysis models have
been developed \cite{Wang-Li-Reddy-2017} such that most implementations
(random survival forests, deep neural networks) can be regarded as black-box
models. Therefore, the problem of the survival analysis result explanation is
topical. However, in contrast to other machine learning models, one of the
main difficulties of the survival model explanation is that the result of most
survival models is a time-dependent function (the survival function, the
hazard function, the cumulative hazard function, etc.). This implies that many
available explanation methods like LIME cannot be applied to survival models.
In order to cope with this difficulty, a new method called SurvLIME (Survival
LIME) for explaining machine learning survival models is proposed, which can
be viewed as an extension or modification of LIME. The main idea of the
proposed method is to apply the Cox proportional hazards model to approximate
the survival model at a local area around a test example. It has been
mentioned that the Cox model considers a linear combination of the example
covariates. Moreover, it is important that the covariates as well as their
combination do not depend on time. Therefore, coefficients of the covariates
can be regarded as quantitative impacts on the prediction. However, we
approximate not a point-valued black-box model prediction, but functions, for
example, the cumulative hazard function (CHF). In accordance with the proposed
explanation method, synthetic examples are randomly generated around the
explainable example, and the CHF is calculated for every synthetic example by
means of the black-box survival model. Simultaneously, we write the CHF
corresponding to the approximating Cox model as a function of the coefficients
of interest. By writing the distance between the CHF provided by the black-box
survival model and the CHF of the approximating Cox model, we construct an
unconstrained convex optimization problem for computing the coefficients of
covariates. Numerical results using synthetic and real data illustrate the
proposed method.

The paper is organized as follows. Related work can be found in Section 2. A
short description of basic concepts of survival analysis, including the Cox
model, is given in Section 3. Basic ideas of the method LIME are briefly
considered in Section 4. Section 5 provides a description of the proposed
SurvLIME and its basic ideas. A formal derivation of the convex optimization
problem for determining important features and a scheme of an algorithm
implementing SurvLIME can be found in Section 6. Numerical experiments with
synthetic data are provided in Section 7. Similar numerical experiments with
real data are given in Section 8. Concluding remarks are provided in Section 9.

\section{Related work}

\textbf{Local explanation methods.} A lot of methods have been developed to
locally explain black-box models. Along with the original LIME
\cite{Ribeiro-etal-2016}, many its modifications have been proposed due to
success and simplicity of the method, for example, ALIME
\cite{Shankaranarayana-Runje-2019}, NormLIME \cite{Ahern-etal-2019}, DLIME
\cite{Zafar-Khan-2019}, Anchor LIME \cite{Ribeiro-etal-2018}, LIME-SUP
\cite{Hu-Chen-Nair-Sudjianto-2018}, LIME-Aleph \cite{Rabold-etal-2019},
GraphLIME \cite{Huang-Yamada-etal-2020}. Another very popular method is the
SHAP \cite{Strumbel-Kononenko-2010} which takes a game-theoretic approach for
optimizing a regression loss function based on Shapley values
\cite{Lundberg-Lee-2017}. Alternative methods are influence functions
\cite{Koh-Liang-2017}, a multiple hypothesis testing framework
\cite{Burns-etal-2019}, and many other methods.

An increasingly important family of methods are based on counterfactual
explanations \cite{Wachter-etal-2017}, which try to explain what to do in
order to achieve a desired outcome by means of finding changes to some
features of an explainable input example such that the resulting data point
called counterfactual has a different specified prediction than the original
input. Due to intuitive and human-friendly explanations provided by this
family of methods, it is extended very quickly
\cite{Goyal-etal-2018,Hendricks-etal-2018,Looveren-Klaise-2019,Waa-etal-2018}.
Counterfactual modifications of LIME have been also proposed by Ramon et al.
\cite{Ramon-etal-2020} and White and Garcez \cite{White-Garcez-2020}.

Many aforementioned explanation methods starting from LIME
\cite{Ribeiro-etal-2016} are based on perturbation techniques
\cite{Fong-Vedaldi-2019,Fong-Vedaldi-2017,Petsiuk-etal-2018,Vu-etal-2019}.
These methods assume that contribution of a feature can be determined by
measuring how prediction score changes when the feature is altered
\cite{Du-Liu-Hu-2019}. One of the advantages of perturbation techniques is
that they can be applied to a black-box model without any need to access the
internal structure of the model. A possible disadvantage of perturbation
technique is its computational complexity when perturbed input examples are of
the high dimensionality.

Descriptions of many explanation methods and various approaches, their
critical reviews can be found in survey papers
\cite{Adadi-Berrada-2018,Arrieta-etal-2019,Carvalho-etal-2019,Guidotti-2019,Rudin-2019}%
.

It should be noted that most explanation methods deal with the point-valued
predictions produced by black-box models. We mean under point-valued
predictions some finite set of possible model outcomes, for example, classes
of examples. A main problem of the considered survival models is that their
outcome is a function. Therefore, we try to propose a new approach, which uses
LIME as a possible tool for its implementing, dealing with CHFs as the model outcomes.

\textbf{Machine learning models in survival analysis}. A review of survival
analysis methods is presented by Wang et al. \cite{Wang-Li-Reddy-2017}. The
Cox model is a very powerful and popular method for dealing with survival
data. Therefore, a lot of approaches modifying the Cox model have been
proposed last decades. In particular, Tibshirani \cite{Tibshirani-1997}
proposed a modification of the Cox model based on the Lasso method in order to
take into account a high dimensionality of survival data. Following this
paper, several modifications of the Lasso methods for the Cox model were
introduced
\cite{Kaneko-etal-2015,Kim-etal-2012,Ternes-etal-2016,Witten-Tibshirani-2010,Zhang-Lu-2007}%
. In order to relax the linear relationship assumption accepted in the Cox
model, a simple neural network as well as the deep neural networks were
proposed by several authors
\cite{Faraggi-Simon-1995,Haarburger-etal-2018,Katzman-etal-2018,Ranganath-etal-2016,Zhu-Yao-Huang-2016}%
. The SVM approach to survival analysis has been also studied by several
authors
\cite{Van_Belle-etal-2011,Khan-Zubek-2008,Shivaswamy-etal-2007,Widodo-Yang-2011}%
. It turned out that the random survival forests (RSFs) became a very
powerful, efficient and popular tool for the survival analysis. Therefore,
this tool and its modifications, which can be regarded as extensions of the
standard random forest \cite{Breiman-2001} on survival data, were proposed and
investigated in many papers, for example, in
\cite{Ibrahim-etal-2008,Mogensen-etal-2012,Nasejje-etal-2017,Omurlu-etal-2009,Schmid-etal-2016,Wang-Zhou-2017,Wright-etal-2016,Wright-etal-2017}%
, in order to take into account the limited survival data.

Most of the above models dealing with survival data can be regarded as
black-box models and should be explained. However, only the Cox model has a
simple explanation due to its linear relationship between covariates.
Therefore, it can be used to approximate more powerful models, including
survival deep neural networks and RSFs, in order to explain predictions of
these models.

\section{Some elements of survival analysis}

\subsection{Basic concepts}

In survival analysis, an example (patient) $i$ is represented by a triplet
$(\mathbf{x}_{i},\delta_{i},T_{i})$, where $\mathbf{x}_{i}^{\mathrm{T}%
}=(x_{i1},...,x_{id})$ is the vector of the patient parameters
(characteristics) or the vector of the example features; $T_{i}$ is time to
event of the example. If the event of interest is observed, then $T_{i}$
corresponds to the time between baseline time and the time of event happening,
in this case $\delta_{i}=1$, and we have an uncensored observation. If the
example event is not observed and its time to event is greater than the
observation time, then $T_{i}$ corresponds to the time between baseline time
and end of the observation, and the event indicator is $\delta_{i}=0$, and we
have a censored observation. Suppose a training set $D$ consists of $n$
triplets $(\mathbf{x}_{i},\delta_{i},T_{i})$, $i=1,...,n$. The goal of
survival analysis is to estimate the time to the event of interest $T$ for a
new example (patient) with feature vector denoted by $\mathbf{x}$ by using the
training set $D$.

The survival and hazard functions are key concepts in survival analysis for
describing the distribution of event times. The survival function denoted by
$S(t)$ as a function of time $t$ is the probability of surviving up to that
time, i.e., $S(t)=\Pr\{T>t\}$. The hazard function $h(t)$ is the rate of event
at time $t$ given that no event occurred before time $t$, i.e.,
\begin{equation}
h(t)=\lim_{\Delta t\rightarrow0}\frac{\Pr\{t\leq T\leq t+\Delta t|T\geq
t\}}{\Delta t}=\frac{f(t)}{S(t)},
\end{equation}
where $f(t)$ is the density function of the event of interest.

By using the fact that the density function can be expressed through the
survival function as
\begin{equation}
f(t)=-\frac{\mathrm{d}S(t)}{\mathrm{d}t},
\end{equation}
we can write the following expression for the hazard rate:
\begin{equation}
h(t)=-\frac{\mathrm{d}}{\mathrm{d}t}\ln S(t).
\end{equation}

Another important concept in survival analysis is the CHF $H(t)$, which is
defined as the integral of the hazard function $h(t)$ and can be interpreted
as the probability of an event at time $t$ given survival until time $t$,
i.e.,
\begin{equation}
H(t)=\int_{-\infty}^{t}h(x)dx.
\end{equation}

The survival function is determined through the hazard function and through
the CHF as follows:
\begin{equation}
S(t)=\exp\left(  -\int_{0}^{t}h(z)\mathrm{d}z\right)  =\exp\left(
-H(t)\right)  .
\end{equation}

The dependence of the above functions on a feature vector $\mathbf{x}$ is
omitted for short.

To compare survival models, the C-index proposed by Harrell et al.
\cite{Harrell-etal-1982} is used. It estimates how good a survival model is at
ranking survival times. It estimates the probability that, in a randomly
selected pair of examples, the example that fails first had a worst predicted
outcome. In fact, this is the probability that the event times of a pair of
examples are correctly ranking.

\subsection{The Cox model}

Let us consider main concepts of the Cox proportional hazards model,
\cite{Hosmer-Lemeshow-May-2008}. According to the model, the hazard function
at time $t$ given predictor values $\mathbf{x}$ is defined as
\begin{equation}
h(t|\mathbf{x},\mathbf{b})=h_{0}(t)\Psi(\mathbf{x},\mathbf{b})=h_{0}%
(t)\exp\left(  \psi(\mathbf{x},\mathbf{b})\right)  . \label{SurvLIME1_10}%
\end{equation}

Here $h_{0}(t)$ is a baseline hazard function which does not depend on the
vector $\mathbf{x}$ and the vector $\mathbf{b}$; $\Psi(\mathbf{x},\mathbf{b})$
is the covariate effect or the risk function; $\mathbf{b}^{\mathrm{T}}%
=(b_{1},...,b_{m})$ is an unknown vector of regression coefficients or
parameters. It can be seen from the above expression for the hazard function
that the reparametrization $\Psi(\mathbf{x},\mathbf{b})=\exp\left(
\psi(\mathbf{x},\mathbf{b})\right)  $ is used in the Cox model. The function
$\psi(\mathbf{x},\mathbf{b})$ in the model is linear, i.e.,
\begin{equation}
\psi(\mathbf{x},\mathbf{b})=\mathbf{b}^{\mathrm{T}}\mathbf{x}=\sum
\nolimits_{k=1}^{m}b_{k}x_{k}.
\end{equation}

In the framework of the Cox model, the survival function $S(t|\mathbf{x}%
,\mathbf{b})$ is computed as
\begin{equation}
S(t|\mathbf{x},\mathbf{b})=\exp(-H_{0}(t)\exp\left(  \psi(\mathbf{x}%
,\mathbf{b})\right)  )=\left(  S_{0}(t)\right)  ^{\exp\left(  \psi
(\mathbf{x},\mathbf{b})\right)  }.
\end{equation}

Here $H_{0}(t)$ is the cumulative baseline hazard function; $S_{0}(t)$ is the
baseline survival function. It is important to note that functions $H_{0}(t)$
and $S_{0}(t)$ do not depend on $\mathbf{x}$ and $\mathbf{b}$.

The partial likelihood in this case is defined as follows:
\begin{equation}
L(\mathbf{b})=\prod_{j=1}^{n}\left[  \frac{\exp(\psi(\mathbf{x}_{j}%
,\mathbf{b}))}{\sum_{i\in R_{j}}\exp(\psi(\mathbf{x}_{i},\mathbf{b}))}\right]
^{\delta_{j}}.
\end{equation}

Here $R_{j}$ is the set of patients who are at risk at time $t_{j}$. The term
\textquotedblleft at risk at time $t$\textquotedblright\ means patients who
die at time $t$ or later.

\section{LIME}

Before studying the LIME modification for survival data, this method is
briefly considered below.

LIME proposes to approximate a black-box model denoted as $f$ with a simple
function $g$ in the vicinity of the point of interest $\mathbf{x}$, whose
prediction by means of $f$ has to be explained, under condition that the
approximation function $g$ belongs to a set of explanation models $G$, for
example, linear models. In order to construct the function $g$ in accordance
with LIME, a new dataset consisting of perturbed samples is generated, and
predictions corresponding to the perturbed samples are obtained by means of
the explained model. New samples are assigned by weights $w_{\mathbf{x}}$ in
accordance with their proximity to the point of interest $\mathbf{x}$ by using
a distance metric, for example, the Euclidean distance.

An explanation (local surrogate) model is trained on new generated samples by
solving the following optimization problem:
\begin{equation}
\arg\min_{g\in G}L(f,g,w_{\mathbf{x}})+\Phi(g).
\end{equation}

Here $L$ is a loss function, for example, mean squared error, which measures
how the explanation is close to the prediction of the black-box model;
$\Phi(g)$ is the model complexity.

A local linear model is the result of the original LIME. As a result, the
prediction is explained by analyzing coefficients of the local linear model.

\section{A basic sketch of SurvLIME}

Suppose that there are available a training set $D$ and an black-box model.
For every new example $\mathbf{x}$, the black-box model with input
$\mathbf{x}$ produces the corresponding output in the form of the CHF
$H(t|\mathbf{x})$ or the hazard function $h(t|\mathbf{x})$. The basic idea
behind the explanation algorithm SurvLIME is to approximate the output of the
black-box model in the form of the CHF by means of the CHF produced by the
Cox model for the same input example $\mathbf{x}$. The idea stems from the
fact that the function $\psi(\mathbf{x},\mathbf{b})$ in the Cox model (see
(\ref{SurvLIME1_10})) is linear and does not depend on time $t$. The linearity
means that coefficients $b_{i}$ of the covariates in $\psi(\mathbf{x}%
,\mathbf{b})$ can be regarded as quantitative impacts on the prediction.
Hence, the largest coefficients indicate the corresponding importance
features. The independence of $\psi(\mathbf{x},\mathbf{b})$ on time $t$ means
that time and covariates can be considered separately, and the optimization
problem minimizing the difference between CHFs is significantly simplifies.

\begin{figure}
[ptb]
\begin{center}
\includegraphics[
height=3.7377in,
width=4.6492in
]%
{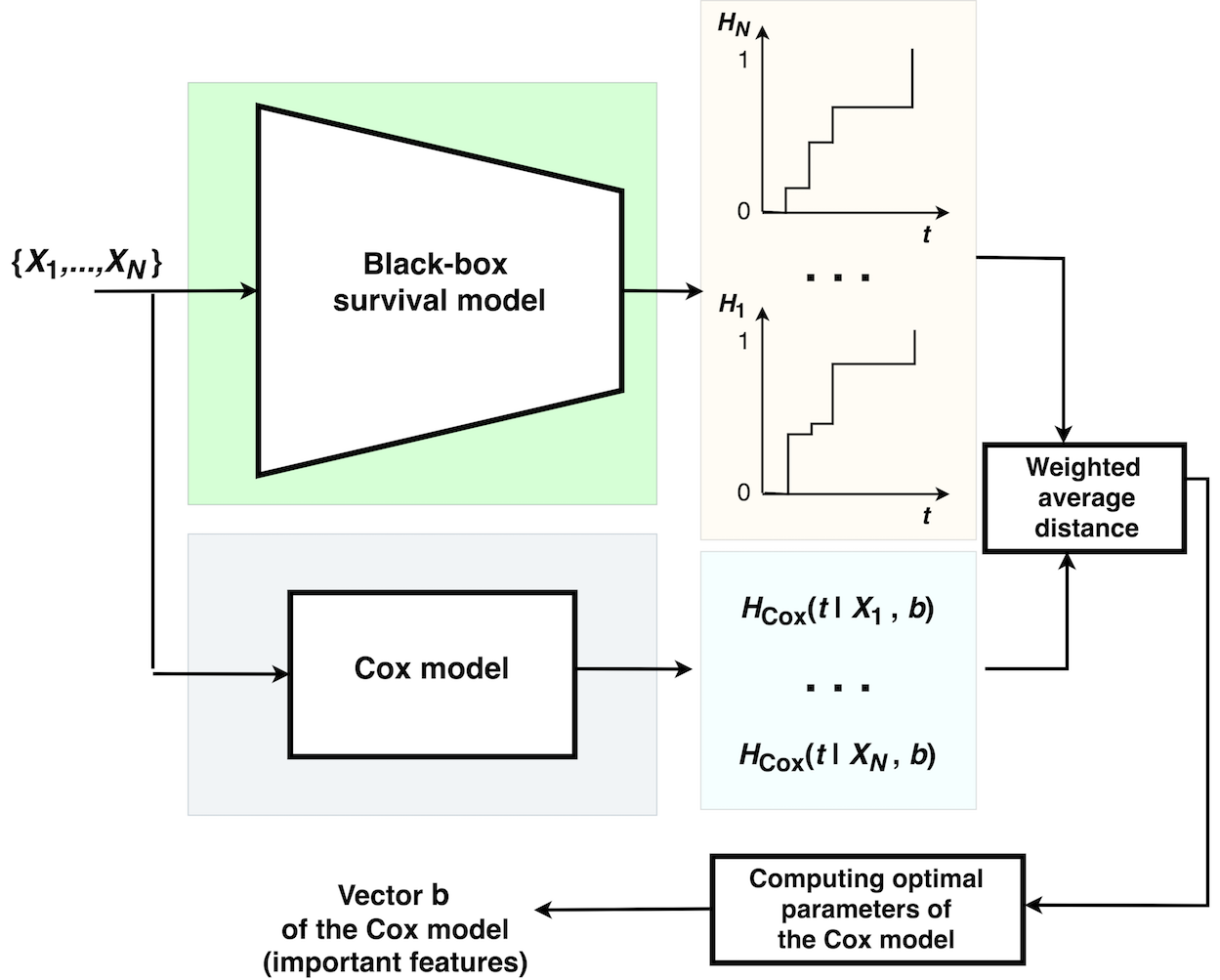}%
\caption{A schematic illustration of the explanation algorithm}%
\label{fig:chf_explain}%
\end{center}
\end{figure}

In order to find the important features of $\mathbf{x}$, we have to compute
some optimal values of elements of vector $\mathbf{b}$ (see the previous
section) of the Cox model such that $H(t|\mathbf{x})$ would be as close as
possible to the Cox CHF denoted as $H_{\text{Cox}}(t|\mathbf{x},\mathbf{b})$.
However, the use of a single point may lead to incorrect results. Therefore,
we generate a lot of nearest points $\mathbf{x}_{k}$ in a local area around
$\mathbf{x}$. For every generated point $\mathbf{x}_{k}$, the CHF
$H(t|\mathbf{x}_{k})$ of the black-box model can be computed as a prediction
provided by the survival black-box model, and the Cox CHF
$H_{\text{Cox}}(t|\mathbf{x}_{k},\mathbf{b})$ can be written as a function of
unknown vector $\mathbf{b}$. Now the optimal values of $\mathbf{b}$ can be
computed by minimizing the average distance between every pair of CHFs
$H(t|\mathbf{x}_{k})$ and $H_{\text{Cox}}(t|\mathbf{x}_{k},\mathbf{b})$ over
all generated points $\mathbf{x}_{k}$. Every distance between CHFs has a
weight $w_{k}$ which depends on the distance between $\mathbf{x}_{k}$ and
$\mathbf{x}$. Smaller distances between $\mathbf{x}_{k}$ and $\mathbf{x}$
produce larger weights of distances between CHFs. If explanation methods like
LIME deal with point-valued predictions of the example processing through the
black-box model, then the proposed method is based on functions
characterizing every point from $D$ or new points $\mathbf{x}$. Fig.
\ref{fig:chf_explain} illustrates the explanation algorithm. It can be seen
from Fig. \ref{fig:chf_explain} that a set of examples $\{\mathbf{x}%
_{1},...,\mathbf{x}_{N}\}$ are fed to the black-box survival model, which
produces a set of CHFs $\{H(t|\mathbf{x}_{1}),...,H(t|\mathbf{x}_{N})\}$.
Simultaneously, we write CHFs $H_{\text{Cox}}(t|\mathbf{x}_{k},\mathbf{b})$,
$k=1,...,N$, as functions of coefficients $\mathbf{b}$ for all generated
examples. The weighted average distance between the CHFs of the Cox model and
the black-box survival model allows us to construct an optimization problem
and to compute the optimal vector $\mathbf{b}$ by solving the optimization problem.%

It should be noted that the above description is only a sketch where every
step is a separate task. All steps will be considered in detail below.

\section{Minimization of distances between functions}

It has been mentioned that the main peculiarity of machine learning survival
models is that the output of models is a function (the CHF or the survival
function). Therefore, in order to approximate the output of the black-box
model by means of the CHF produced by the Cox model at the input example
$\mathbf{x}$, we have to generate many points $\mathbf{x}_{k}$ in a local area
around $\mathbf{x}$ and to consider the mean distance between the CHFs for
generated points $\mathbf{x}_{k}$, $k=1,...,N$, and the Cox model CHF for
point $\mathbf{x}$. Before deriving the mean distance and its minimizing, we
introduce some notations and conditions.

Let $t_{0}<t_{1}<...<t_{m}$ be the distinct times to event of interest, for
example, times to deaths from the set $\{T_{1},...,T_{n}\}$, where $t_{0}%
=\min_{k=1,...,n}T_{k}$ and $t_{m}=\max_{k=1,...,n}T_{k}$. The black-box
model maps the feature vectors $\mathbf{x}\in\mathbb{R}^{d}$ into piecewise
constant CHFs $H(t|\mathbf{x})$ having the following properties:

\begin{enumerate}
\item $H(t|\mathbf{x})\geq0$ for all $t$

\item $\max_{t}H(t|\mathbf{x})<\infty$

\item $\int_{0}^{\infty}H(t|\mathbf{x})\mathrm{d}t\rightarrow\infty$
\end{enumerate}

Let us introduce the time $T=t_{m}+\gamma$ in order to restrict the integral
of $H(t|\mathbf{x})$, where $\gamma$ is a very small positive number. Let
$\Omega=[0,T]$. Then we can write
\begin{equation}
\int_{\Omega}H(t|\mathbf{x})\mathrm{d}t<\infty.
\end{equation}

Since the CHF $H(t|\mathbf{x})$ is piecewise constant, then it can be written
in a special form. Let us divide the set $\Omega$ into $m+1$ subsets
$\Omega_{0},...,\Omega_{m}$ such that

\begin{enumerate}
\item $\Omega=\cup_{j=0,...,m}\Omega_{j}$

\item $\Omega_{m}=[t_{m},T]$, $\Omega_{j}=[t_{j},t_{j+1})$, $\forall
j\in\{0,...,m-1\}$

\item $\Omega_{j}\cap\Omega_{k}=\emptyset$ for $\forall j\neq k$
\end{enumerate}

Let us introduce the indicator functions
\begin{equation}
\chi_{\Omega_{j}}(t)=\left\{
\begin{array}
[c]{cc}%
1, & t\in\Omega_{j},\\
0, & t\notin\Omega_{j}.
\end{array}
\right.
\end{equation}

Hence, the CHF $H(t|\mathbf{x})$ can be expressed through the indicator
functions as follows:
\begin{equation}
H(t|\mathbf{x})=\sum_{j=0}^{m}H_{j}(\mathbf{x})\cdot\chi_{\Omega_{j}}(t)
\end{equation}
under additional condition $H_{j}(\mathbf{x})\geq\varepsilon>0$, where
$\varepsilon$ is a small positive number. Here $H_{j}(\mathbf{x})$ is a part
of the CHF in interval $\Omega_{j}$. It is important that $H_{j}(\mathbf{x})$
does not depend on $t$ and is constant in interval $\Omega_{j}$. The last
condition will be necessary below in order to deal with logarithms of the CHFs.

Let $g$ be a monotone function. Then there holds
\begin{equation}
g(H(t|\mathbf{x}))=\sum_{j=0}^{m}g(H_{j}(\mathbf{x}))\chi_{\Omega_{j}}(t).
\end{equation}

By using the above representation of the CHF and integrating it over $\Omega$,
we get
\begin{align}
\int_{\Omega}H(t|\mathbf{x})\mathrm{d}t  &  =\int_{\Omega}\left[  \sum
_{j=0}^{m}H_{j}(\mathbf{x})\chi_{\Omega_{j}}(t)\right]  \mathrm{d}t\nonumber\\
&  =\sum_{j=0}^{m}H_{j}(\mathbf{x})\left[  \int_{\Omega}\chi_{\Omega_{j}%
}(t)\right]  \mathrm{d}t=\sum_{j=0}^{m}H_{j}(\mathbf{x})\left(  t_{j+1}%
-t_{j}\right)  .
\end{align}

The same expressions can be written for the Cox CHF:
\begin{equation}
H_{\text{Cox}}(t|\mathbf{x},\mathbf{b})=H_{0}(t)\exp\left(  \mathbf{b}%
^{\mathrm{T}}\mathbf{x}\right)  =\sum_{j=0}^{m}\left[  H_{0j}\exp\left(
\mathbf{b}^{\mathrm{T}}\mathbf{x}\right)  \right]  \chi_{\Omega_{j}%
}(t),\ H_{0j}\geq\varepsilon.
\end{equation}

It should be noted that the distance between two CHFs can be replaced with the
distance between two logarithms of the corresponding CHFs for the optimization
problem. The introduced condition $H_{j}(\mathbf{x})\geq\varepsilon>0$ allows
to use logarithms. Therefore, in order to get the convex optimization problem
for finding optimal values of $\mathbf{b}$, we consider logarithms of the
CHFs. It is important to point out that the difference of logarithms of the
CHFs is not equal to the difference between CHFs themselves. However, we make
this replacement to simplify the optimization problem for computing important features.

Let $\phi(t|\mathbf{x}_{k})$ and $\phi_{\text{Cox}}(t|\mathbf{x}%
_{k},\mathbf{b})$ be logarithms of $H(t|\mathbf{x}_{k})$ and $H_{\text{Cox}%
}(t|\mathbf{x}_{k},\mathbf{b})$. Here $\mathbf{x}_{k}$ is a generated point.
The difference between functions $\phi(t|\mathbf{x}_{k})$ and $\phi
_{\text{Cox}}(t|\mathbf{x}_{k},\mathbf{b})$ can be written as follows:
\begin{align}
&  \phi(t|\mathbf{x}_{k})-\phi_{\text{Cox}}(t|\mathbf{x}_{k},\mathbf{b}%
)\nonumber\\
&  =\sum_{j=0}^{m}(\ln H_{j}(\mathbf{x}_{k}))\chi_{\Omega_{j}}(t)-\sum
_{j=0}^{m}\left(  \ln(H_{0j}\exp\left(  \mathbf{b}^{\mathrm{T}}\mathbf{x}%
_{k}\right)  )\right)  \chi_{\Omega_{j}}(t)\nonumber\\
&  =\sum_{j=0}^{m}\left(  \ln H_{j}(\mathbf{x}_{k})-\ln H_{0j}-\mathbf{b}%
^{\mathrm{T}}\mathbf{x}_{k}\right)  \chi_{\Omega_{j}}(t).
\end{align}

Let us consider the distance between functions $\phi(t|\mathbf{x}_{k})$ and
$\phi_{\text{Cox}}(t|\mathbf{x}_{k},\mathbf{b})$ in metric $L_{2}$:%
\begin{align}
D_{2,k}\left(  \phi,\phi_{\text{Cox}}\right)   &  =\left\Vert \phi
(t|\mathbf{x}_{k})-\phi_{\text{Cox}}(t|\mathbf{x}_{k},\mathbf{b})\right\Vert
_{2}^{2}\nonumber\\
&  =\int_{\Omega}\left\vert \phi(t|\mathbf{x}_{k})-\phi_{\text{Cox}%
}(t|\mathbf{x}_{k},\mathbf{b})\right\vert ^{2}\mathrm{d}t\nonumber\\
&  =\sum_{j=0}^{m}\left(  \ln H_{j}(\mathbf{x}_{k})-\ln H_{0j}-\mathbf{b}%
^{\mathrm{T}}\mathbf{x}_{k}\right)  ^{2}\left(  t_{j+1}-t_{j}\right)  .
\end{align}

The function $\ln H_{j}(\mathbf{x}_{k})-\ln H_{0j}-\mathbf{b}^{\mathrm{T}%
}\mathbf{x}_{k}$ is linear with $\mathbf{b}$ and, therefore, convex. Moreover,
it is easy to prove by taking the second derivative over $b_{j}$ that the
function $\left(  A-\mathbf{b}^{\mathrm{T}}\mathbf{x}_{k}\right)  ^{2}$ is
also convex, where $A=\ln H_{j}(\mathbf{x}_{k})-\ln H_{0j}$. Since the term
$\left(  t_{j+1}-t_{j}\right)  $ is positive, then the function $D_{2,k}%
\left(  \phi,\phi_{\text{Cox}}\right)  $ is also convex as a linear
combination of convex functions with positive coefficients.

According to the explanation algorithm, we have to consider many points
$\mathbf{x}_{k}$ generated in a local area around $\mathbf{x}$ and to minimize
the objective function $\sum_{k=1}^{N}w_{k}D_{2,k}\left(  \phi,\phi
_{\text{Cox}}\right)  $, which takes into account all these generated examples
and the corresponding weights of the examples. It is obvious that this
function is also convex with respect to $\mathbf{b}$. The weight can be
assigned to point $\mathbf{x}_{k}$ as a decreasing function of the distance
between $\mathbf{x}$ and $\mathbf{x}_{k}$, for example, $w_{k}=K(\mathbf{x}%
,\mathbf{x}_{k})$, where $K(\cdot,\cdot)$ is a kernel. In numerical
experiments, we use the function defined in (\ref{SurvLIME_60}).

Finally, we write the following convex optimization problem:
\begin{equation}
\min_{\mathbf{b}}\sum_{k=1}^{N}w_{k}\sum_{j=0}^{m}\left(  \ln H_{j}%
(\mathbf{x}_{k})-\ln H_{0j}-\mathbf{b}^{\mathrm{T}}\mathbf{x}_{k}\right)
^{2}\left(  t_{j+1}-t_{j}\right)  .
\end{equation}

One of the difficulties of solving the above problem is that the difference
between functions $H(t|\mathbf{x})$ and $H_{\text{Cox}}(t|\mathbf{x}%
,\mathbf{b})$ may be significantly different from the distance between their
logarithms. Therefore, in order to take into account this fact, it is proposed
to introduce weights which \textquotedblleft straighten\textquotedblright%
\ functions $\phi(t|\mathbf{x})$ and $\phi_{\text{Cox}}(t|\mathbf{x}%
,\mathbf{b})$. These weights are defined as
\begin{equation}
v(t|\mathbf{x})=\frac{H(t|\mathbf{x})}{\phi(t|\mathbf{x})}=\frac
{H(t|\mathbf{x})}{\ln(H(t|\mathbf{x}))}.
\end{equation}

Taking into account these weights and their representation $v_{kj}=$
$H_{j}(\mathbf{x}_{k})/\ln\left(  H_{j}(\mathbf{x}_{k})\right)  $, the
optimization problem can be rewritten as
\begin{equation}
\min_{\mathbf{b}}\sum_{k=1}^{N}w_{k}\sum_{j=0}^{m}v_{kj}^{2}\left(  \ln
H_{j}(\mathbf{x}_{k})-\ln H_{0j}-\mathbf{b}^{\mathrm{T}}\mathbf{x}_{k}\right)
^{2}\left(  t_{j+1}-t_{j}\right)  . \label{SurvLIME1_40}%
\end{equation}

The problem can be solved by many well-known methods of the convex programming.

Finally, we write the following scheme of Algorithm \ref{alg:SurvLIME1}.

\begin{algorithm}
\caption{The algorithm for computing vector $\bf {b}$ for point $\bf {x}$ }\label{alg:SurvLIME1}%
\begin{algorithmic}
[1]\REQUIRE Training set $D$; point of interest $\mathbf{x}$; the number of
generated points $N$; the black-box survival model for explaining
$f(\mathbf{x})$
\ENSURE Vector $\mathbf{b}$ of important features
\STATE Compute the baseline cumulative hazard function $H_{0}(t)$ of the
approximating Cox model on dataset $D$ by using the Nelson--Aalen estimator
\STATE Generate $N-1$ random nearest points $\mathbf{x}_{k}$ in a local area
around $\mathbf{x}$, point $\mathbf{x}$ is the $N$-th point
\STATE Get the prediction of the cumulative hazard function $H(t|\mathbf{x}%
_{k})$ by using the black-box survival model (the function $f$)
\STATE Compute weights $w_{k}=K(\mathbf{x},\mathbf{x}_{k})$ of perturbed
points, $k=1,...,N$
\STATE Compute weights $v_{kj}=$ $H_{j}(\mathbf{x}_{k})/\ln\left(
H_{j}(\mathbf{x}_{k})\right)  $, $k=1,...,N$, $j=0,...,m$
\STATE Find vector $\mathbf{b}$ by solving the convex optimization problem
(\ref{SurvLIME1_40})
\end{algorithmic}
\end{algorithm}

\section{Numerical experiments with synthetic data}

In order to study the proposed explanation algorithm, we generate random
survival times to events by using the Cox model estimates. This generation
allows us to compare initial data for generating every points and results of SurvLIME.

\subsection{Generation of random covariates, survival times and perturbations}

Two clusters of covariates $\mathbf{x}\in\mathbb{R}^{d}$ are randomly
generated such that points of every cluster are generated from the uniform
distribution in a sphere. The covariate vectors are generated in the
$d$-sphere with some predefined radius $R$. The center $p$ of the $d$-sphere
and its radius $R$ are parameters of experiments. There are several methods
for the uniform sampling of points $\mathbf{x}$ in the $d$-sphere with the
unit radius $R=1$, for example, \cite{Barthe-etal-2005,Harman-Lacko-2010}.
Then every generated point is multiplied by $R$. The following parameters for
points of every cluster are used:

\begin{enumerate}
\item cluster 0: center $p_{0}=(0,0,0,0,0)$; radius $R=8$; number of points
$N=1000$;

\item cluster 1: center $p_{1}=(4,-8,2,4,2)$; radius $R=8$; number of points
$N=1000$.
\end{enumerate}

The parameters of clusters are chosen in order to get some intersecting area
containing points from both the clusters, in particular, the radius is
determined from the centers as follows:
\begin{equation}
R=\left\lceil \frac{\left\Vert p_{0}-p_{1}\right\Vert _{2}}{2}\right\rceil +2.
\end{equation}

\begin{figure}
[ptb]
\begin{center}
\includegraphics[
height=2.4917in,
width=2.4917in
]%
{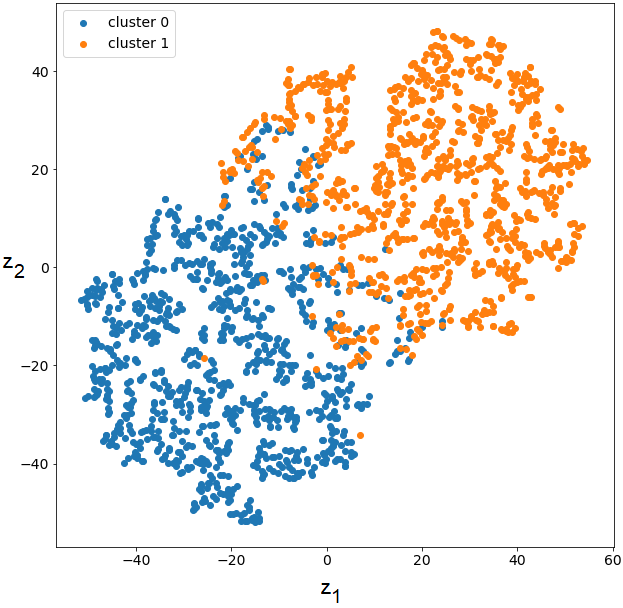}%
\caption{Two clusters of generated covariates depicted by using the t-SNE
method}%
\label{fig:two_cluster_tsne}%
\end{center}
\end{figure}

The clusters after using the well-known t-SNE algorithm are depicted in Fig.
\ref{fig:two_cluster_tsne}.

In order to generate random survival times by using the Cox model estimates,
we apply results obtained by Bender et al. \cite{Bender-etal-2005} for
survival time data for the Cox model with Weibull distributed survival times.
The Weibull distribution with the scale $\lambda$ and shape $v$ parameters is
used to generate appropriate survival times for simulation studies because
this distribution shares the assumption of proportional hazards with the Cox
regression model \cite{Bender-etal-2005}. Taking into account the parameters
$\lambda$ and $v$ of the Weibull distribution, we use the following expression
for generated survival times \cite{Bender-etal-2005}:
\begin{equation}
T=\left(  \frac{-\ln(U)}{\lambda\exp(\mathbf{b}^{\mathrm{T}}\mathbf{x}%
)}\right)  ^{1/v}, \label{SurvLIME1_84}%
\end{equation}
where $U$ is the random variable uniformly distributed in interval $[0,1]$.

Parameters of the generation for every cluster are

\begin{enumerate}
\item cluster 0: $\lambda=10^{-5}$, $v=2$, $\mathbf{b}^{\mathrm{T}}%
=(10^{-6},0.1,-0.15,10^{-6},10^{-6})$;

\item cluster 1: $\lambda=10^{-5}$, $v=2$, $\mathbf{b}^{\mathrm{T}}%
=(10^{-6},-0.15,10^{-6},10^{-6},-0.1)$.
\end{enumerate}

It can be seen from the above that every vector $\mathbf{b}$ has three almost
zero-valued elements and two \textquotedblleft large\textquotedblright%
\ elements which will correspond to important features. Generated values
$T_{i}$ are restricted by the condition: if $T_{i}>2000$, then $T_{i}$ is
replaced with value $2000$. The event indicator $\delta_{i}$ is generated from
the binomial distribution with probabilities $\Pr\{\delta_{i}=1\}=0.9$,
$\Pr\{\delta_{i}=0\}=0.1$.%

Perturbation is one of the steps of the algorithm. According to it, we
generate $N$ nearest points $\mathbf{x}_{k}$ in a local area around
$\mathbf{x}$. These points are uniformly generated in the $d$-sphere with some
predefined radius $r=0.5$ and the center at point $\mathbf{x}$. In numerical
experiments, $N=1000$. Weights to every point are assigned as follows:
\begin{equation}
w_{k}=1-\sqrt{\frac{\left\Vert \mathbf{x}-\mathbf{x}_{k}\right\Vert _{2}}{r}}.
\label{SurvLIME_60}%
\end{equation}

\subsection{Black-box models and approximation measures}

As black-box models, we use the Cox model and the RSF model
\cite{Ibrahim-etal-2008}. The RSF consists of $250$ decision survival trees.
The approximating Cox model has the baseline CHF $H_{0}(t)$ constructed on
generated training data using the Nelson--Aalen estimator. The Cox model is
used in order to check whether the selected important features explaining the
CHF $H(t|\mathbf{x})$ at point $\mathbf{x}$ coincide with the corresponding
features accepted in the Cox model for generating training set. It should be
noted that the Cox model as well as the RSF are viewed as 
black-box models whose predictions (CHFs or survival functions) are explained.
To study how different cases impact on the quality of the approximation, we
use the following two measures for the Cox model:
\begin{equation}
RMSE_{\text{model}}=\sqrt{\frac{1}{n}\sum_{i=1}^{n}\left\Vert \mathbf{b}%
_{i}^{\text{model}}-\mathbf{b}_{i}^{\text{expl}}\right\Vert _{2}},
\end{equation}%
\begin{equation}
RMSE_{\text{true}}=\sqrt{\frac{1}{n}\sum_{i=1}^{n}\left\Vert \mathbf{b}%
_{i}^{\text{true}}-\mathbf{b}_{i}^{\text{expl}}\right\Vert _{2}},
\end{equation}
where $\mathbf{b}_{i}^{\text{model}}$ are coefficients of the Cox model which
is used as the black-box model; $\mathbf{b}_{i}^{\text{true}}$ are
coefficients used for training data generation (see (\ref{SurvLIME1_84}));
$\mathbf{b}_{i}^{\text{expl}}$ are explaining coefficients obtained by using
the proposed algorithm.

The first measure characterizes how the obtained important features coincide
with the corresponding features obtained by using the Cox model as the
black-box model. The second measure considers how the obtained important
feature coincide with the features used for generating the random times to
events. Every measure is calculated by taking randomly $n$ points $\mathbf{x}$
from the testing set and compute the corresponding coefficients.

In order to investigate the quality of explanation when the black-box model
is the RSF, we use another measure:%
\begin{equation}
RMSE_{\text{approx}}=\sqrt{\frac{1}{n}\sum_{i=1}^{n}\sum_{j\in J}\left(
H(t_{j}|\mathbf{x}_{i})-H_{\text{Cox}}\left(  t_{j}|\mathbf{x}_{i}%
,\mathbf{b}_{i}^{\text{expl}}\right)  \right)  ^{2}},
\end{equation}
where $J$ is a set of time indices for computing the measure.

This measure considers how the obtained Cox model approximation $H_{\text{Cox}%
}\left(  t_{j}|\mathbf{x}_{i},\mathbf{b}_{i}^{\text{expl}}\right)  $ is close
to the RSF output $H(t_{j}|\mathbf{x}_{i})$. We cannot estimate the proximity
of important features explaining the model because we do not have the
corresponding features for the RSF. A comparison of the important features
with the generated ones is also incorrect because we explain the model (the
RSF) output ($H(t|\mathbf{x}_{i})$), but not training data. Therefore, we use
the above measure to estimate the proximity of two CHFs: the Cox model
approximation and the RSF output.

\subsection{Experiment 1}

To evaluate the algorithm, 900 examples are randomly selected from every
cluster for training and 100 examples are for testing. Three cases of training
and testing the black-box Cox and RSF models are studied:

\begin{enumerate}
\item cluster 0 for training and for testing;

\item cluster 1 for training and for testing;

\item clusters 0 and 1 are jointly used for training and separately for testing.
\end{enumerate}

The testing phase includes:

\begin{itemize}
\item Computing the explanation vector $\mathbf{b}^{\text{expl}}$ for every
point from the testing set.

\item Depicting the best, mean and worst approximations in accordance with
Euclidean distance between vectorsd $\mathbf{b}^{\text{expl}}$ and
$\mathbf{b}^{\text{model}}$ (for the Cox model) and with Euclidean distance
between $H(t_{j}|\mathbf{x}_{i})$ and $H_{\text{Cox}}\left(  t_{j}%
|\mathbf{x}_{i},\mathbf{b}_{i}^{\text{expl}}\right)  $ (for the RSF). In order
to get these approximations, points with the best, mean and worst
approximations are selected among all testing points.

\item Computing measures $RMSE_{\text{model}}$ and $RMSE_{\text{true}}$ for
the Cox model and $RMSE_{\text{approx}}$ for the RSF over all points of the
testing set.
\end{itemize}

\begin{figure}
[ptb]
\begin{center}
\includegraphics[
height=3.2456in,
width=3.3641in
]%
{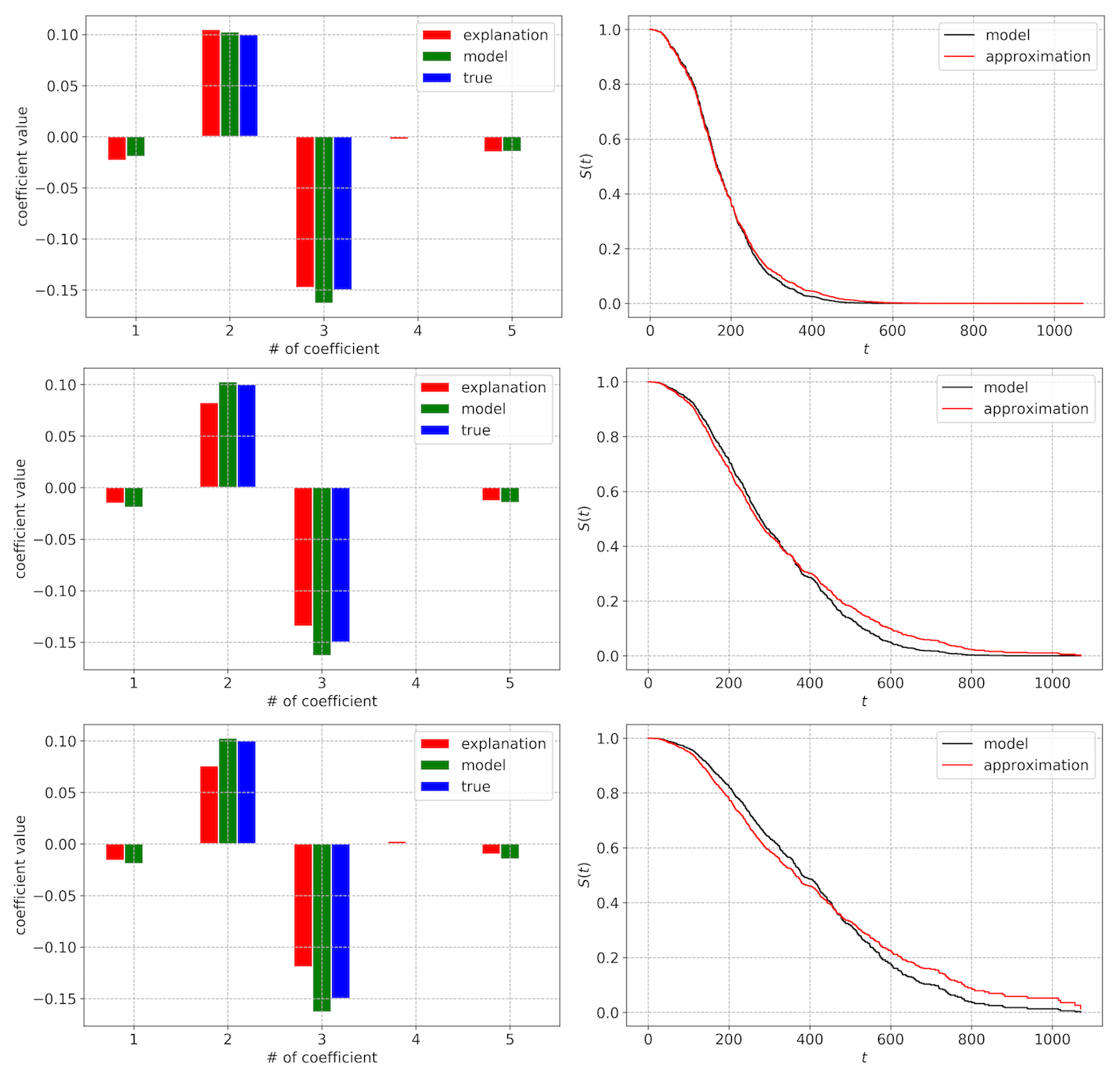}%
\caption{The best, mean and worst approximations for the Cox model (training
and testing sets from cluster 0)}%
\label{cox_ncls100}%
\end{center}
\end{figure}

\begin{figure}
[ptb]
\begin{center}
\includegraphics[
height=3.2949in,
width=3.416in
]%
{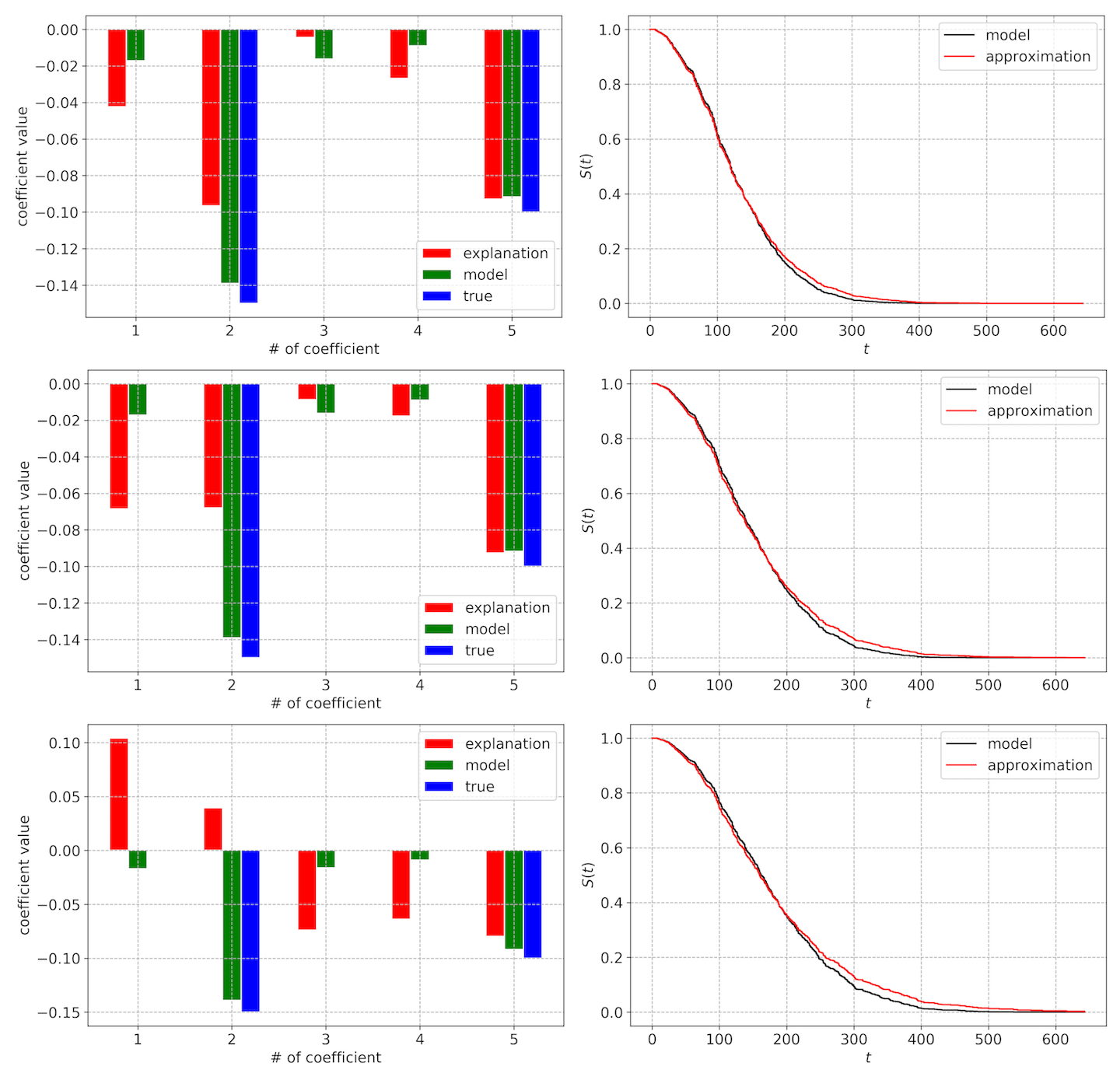}%
\caption{The best, mean and worst approximations for the Cox model (training
and testing sets from cluster 1)}%
\label{cox_ncls111}%
\end{center}
\end{figure}

The three cases (best (pictures in the first row), mean (pictures in the
second row) and worst (pictures in the third row)) of approximations for the
black-box Cox model under condition that cluster 0 is used for training and
testing are depicted in Fig. \ref{cox_ncls100}. Left pictures show values of
important features $\mathbf{b}^{\text{expl}}$, $\mathbf{b}^{\text{model}}$ and
$\mathbf{b}^{\text{true}}$. It can be seen from these pictures that all
experiments show very clear coincidence of important features for all models.
Right pictures in Fig. \ref{cox_ncls100} show survival functions obtained from
the black-box Cox model and from the Cox approximation. It follows from the
pictures that the approximation is perfect even for the worst case. Similar
results can be seen from Fig. \ref{cox_ncls111}, where training and testing
examples are taken from cluster 1.%

%

\begin{figure}
[ptb]
\begin{center}
\includegraphics[
height=3.2128in,
width=3.3183in
]%
{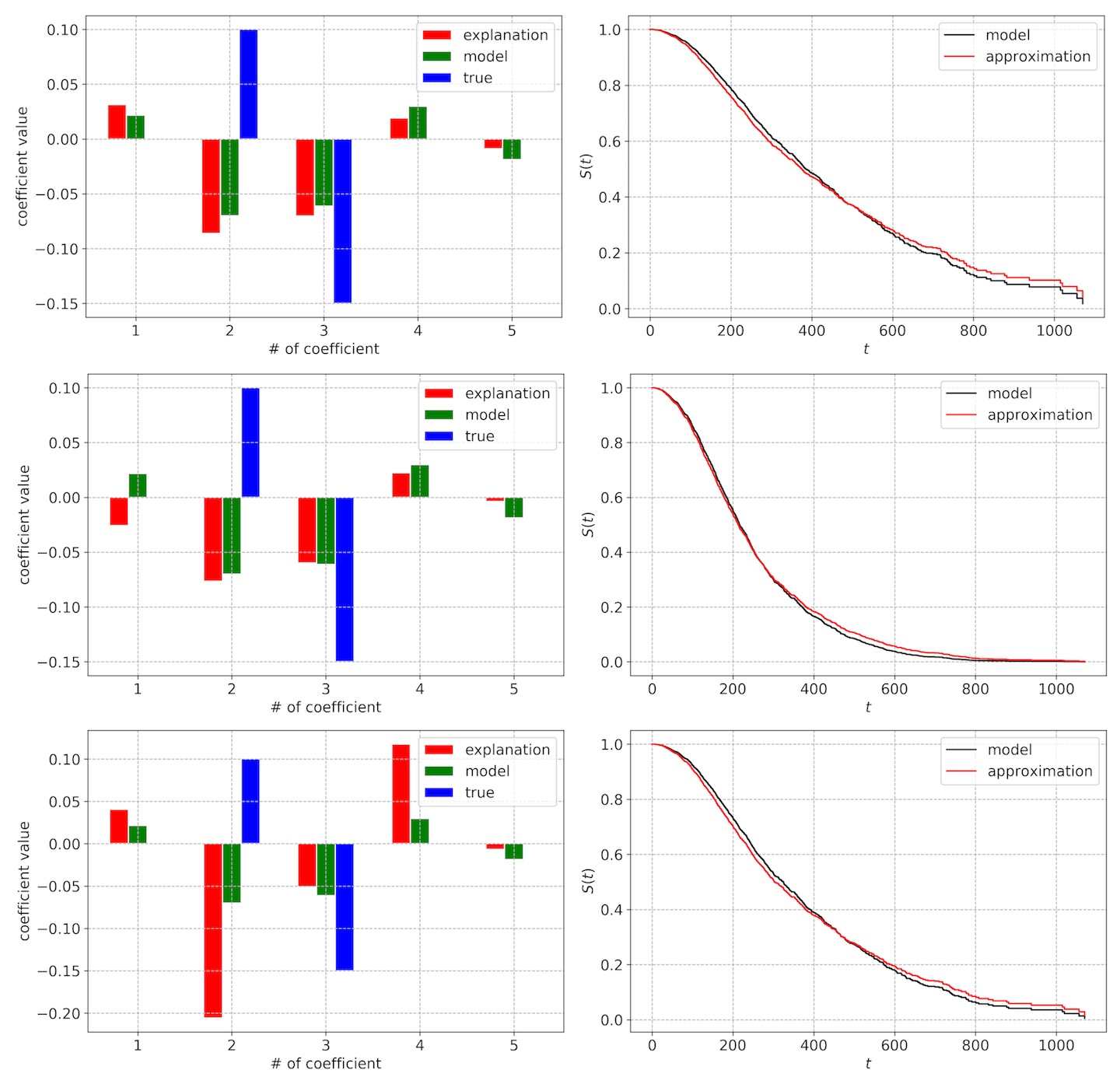}%
\caption{The best, mean and worst approximations for the Cox model (the
training set from clusters 0,1 and the testing set from cluster 0)}%
\label{cox_ncls2010}%
\end{center}
\end{figure}

\begin{figure}
[ptb]
\begin{center}
\includegraphics[
height=3.1799in,
width=3.2837in
]%
{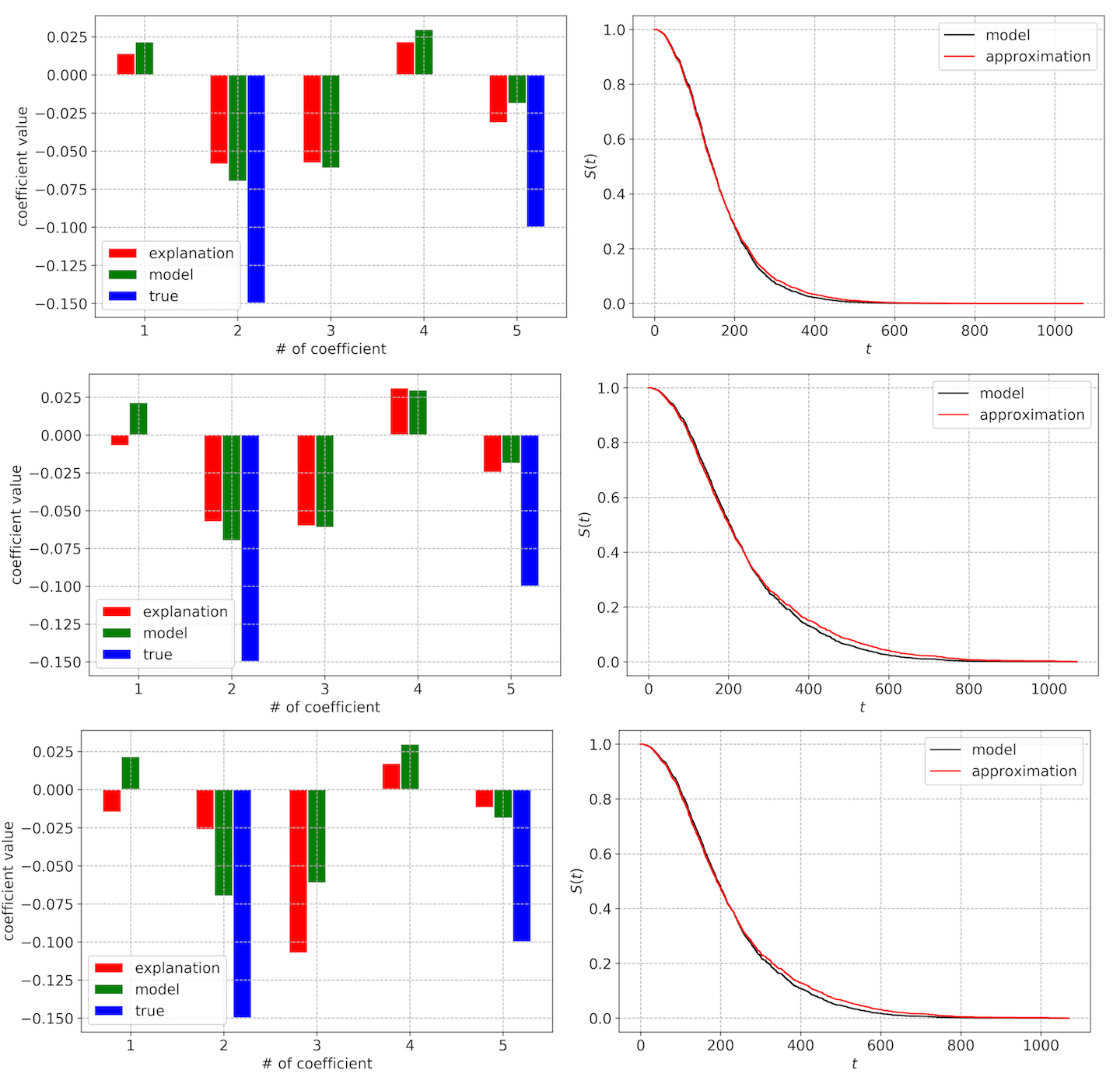}%
\caption{The best, mean and worst approximations for the Cox model (the
training set from clusters 0,1 and the testing set from cluster 1)}%
\label{cox_ncls2011}%
\end{center}
\end{figure}

Figs. \ref{cox_ncls2010} and \ref{cox_ncls2011} illustrate different results
corresponding to cases when training examples are taken from cluster 0 and
cluster 1. One can see that the approximation of survival functions is
perfect, but important features obtained by SurvLIME do not coincide with the
features used for generating random times in accordance with the Cox model.
This fact can be explained in the following way. We try to explain the CHF
obtained by using the black-box model (the Cox model in the considered
case). But the black-box Cox model is trained on all examples from two
different clusters. We have a mix of data from two clusters with different
parameters. Therefore, this model itself provides results different from the
generation model. At the same time, one can observe from Figs.
\ref{cox_ncls2010} and \ref{cox_ncls2011} that the explaining important
features coincide with the features which are obtained from the black-box
Cox model. These results are interesting. They show that we explain CHFs of
the black-box model, but not CHFs of training data.%

%

The approximation accuracy measures for four cases are given in Table
\ref{t:cox_ncls}. In fact, the table repeats the results shown in Figs.
\ref{cox_ncls100}-\ref{cox_ncls2011}.%

\begin{table}[tbp] \centering
\caption{Approximation accuracy measures for four cases of using the black-box Cox model}%
\begin{tabular}
[c]{ccccc}\hline
Clusters for & Cluster for & $RMSE_{\text{model}}$ &
\multicolumn{2}{c}{$RMSE_{\text{true}}$}\\\cline{4-5}%
training & testing &  & $0$ & $1$\\\hline
$0$ & $0$ & $0.016$ & $0.014$ & \\\hline
$1$ & $1$ & $0.038$ &  & $0.048$\\\hline
$\{0,1\}$ & $0$ & $0.023$ & $0.089$ & \\\hline
$\{0,1\}$ & $1$ & $0.014$ &  & $0.065$\\\hline
\end{tabular}
\label{t:cox_ncls}%
\end{table}%

\begin{figure}
[ptb]
\begin{center}
\includegraphics[
height=3.2396in,
width=3.6685in
]%
{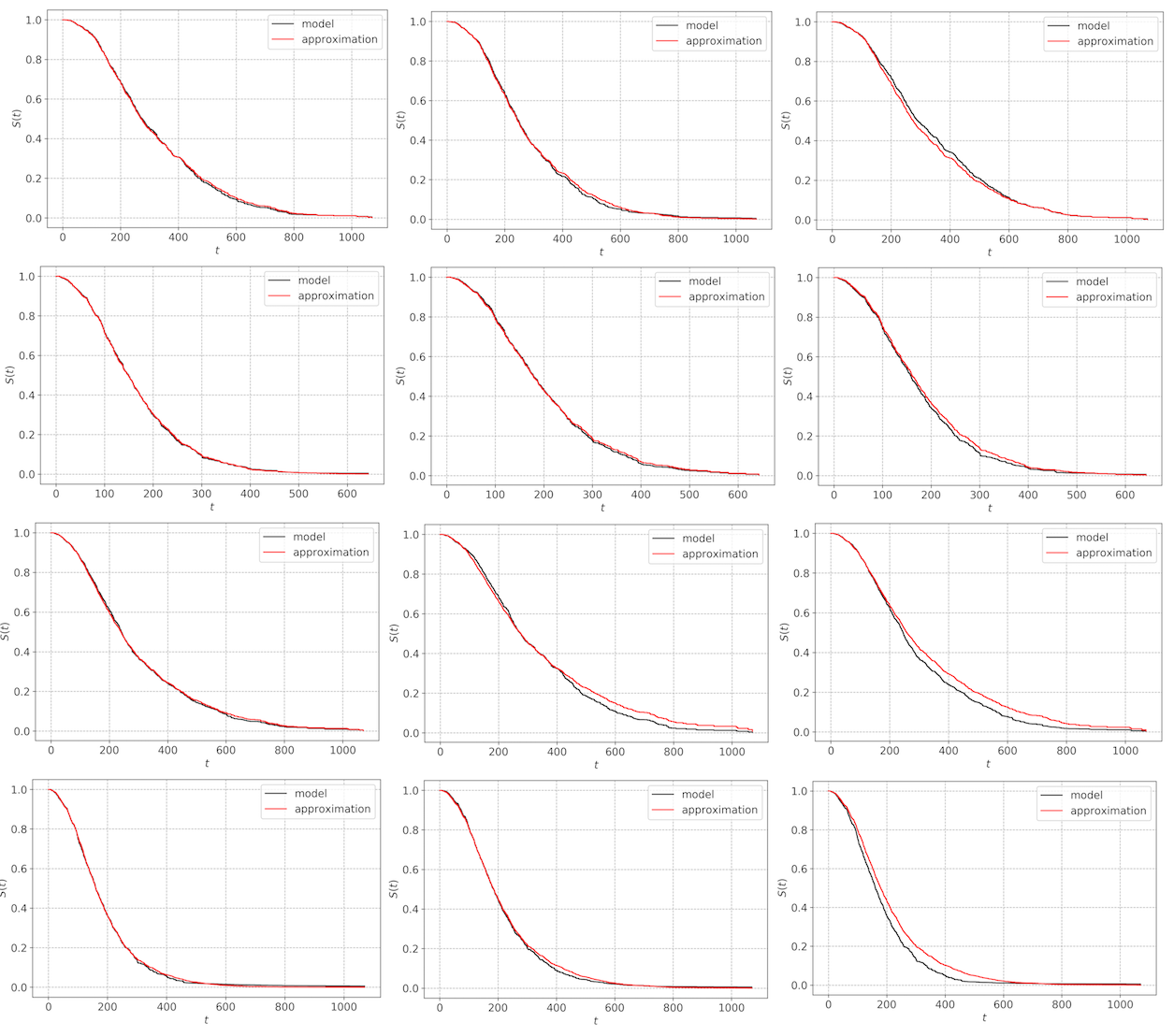}%
\caption{The best, mean and worst approximations for the RSF model}%
\label{rsf_ncls}%
\end{center}
\end{figure}

In the same way, we study the black-box RSF model by using three cases of
its training and testing. The results are shown in Fig. \ref{rsf_ncls} where
the first, second, third, and fourth rows correspond to the four cases of
experiments: cluster 0 for training and for testing; cluster 1 for training
and for testing; clusters 0 and 1 are jointly used for training and separately
for testing. Every row contains pairs of survival functions corresponding to
the best, mean and worst approximations. The important features are not shown
in Fig. \ref{rsf_ncls} because they cannot be compared with the features of
the generative Cox model. Moreover, the RSF does not provide the important
features like the Cox model. One can again see from Fig. \ref{rsf_ncls} that
SurvLIME illustrates the perfect approximation of the RSF output by the Cox model.%

\subsection{Experiment 2}

\begin{figure}
[ptb]
\begin{center}
\includegraphics[
height=4.9026in,
width=3.0614in
]%
{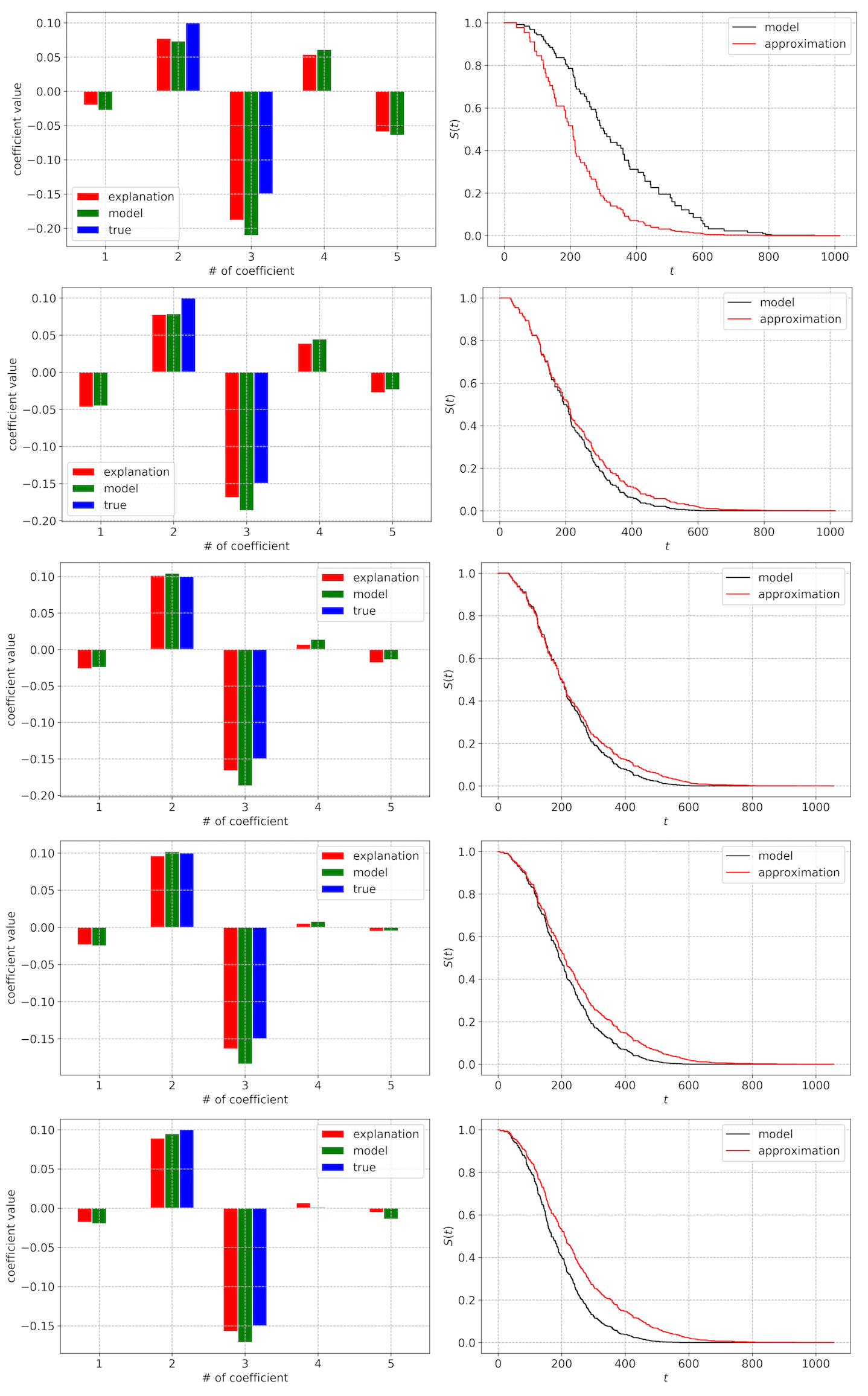}%
\caption{Comparison of important features (left pictures) and survival
functions (right pictures) for the black-box Cox model by $100$, $200$,
$300$, $400$, $500$ training examples}%
\label{cox_sample_size}%
\end{center}
\end{figure}

In the second experiment, our aim is to study how SurvLIME depends on the
number of training examples. We use only the Cox model for explaining which is
trained on $100$, $200$, $300$, $400$, $500$ examples and tested on $100$
examples from cluster 0. We study how the difference between $\mathbf{b}%
^{\text{model}}$ and $\mathbf{b}^{\text{true}}$ depends of the sample size.
Results of experiments are shown in Fig. \ref{cox_sample_size} where rows
correspond to $100$, $200$, $300$, $400$, $500$ training examples,
respectively, left pictures illustrate relationships between important
features, right pictures show survival functions obtained from the black-box
Cox model and from the Cox approximation (see similar pictures in Figs.
\ref{cox_ncls100}-\ref{cox_ncls2011}). It is interesting to observe in Fig.
\ref{cox_sample_size} how the vectors $\mathbf{b}^{\text{expl}}$,
$\mathbf{b}^{\text{model}}$ and $\mathbf{b}^{\text{true}}$ approach each other
with the number of training examples.%

The measures $RMSE_{\text{model}}$ and $RMSE_{\text{true}}$ as functions of
the sample size are provided in Table \ref{t:cox_sample_size}. It can be seen
from Table \ref{t:cox_sample_size} a tendency of the measures to be reduced
with increase of the sample size for training. The C-index as a measure of the
black-box model quality is also given in Table \ref{t:cox_sample_size}.%

\begin{table}[tbp] \centering
\caption{Approximation accuracy measures for four cases of using the black-box Cox model}%
\begin{tabular}
[c]{cccc}\hline
& C-index & $RMSE_{\text{model}}$ & $RMSE_{\text{true}}$\\\hline
$100$ & $0.777$ & $0.027$ & $0.042$\\\hline
$200$ & $0.785$ & $0.018$ & $0.031$\\\hline
$300$ & $0.787$ & $0.021$ & $0.015$\\\hline
$400$ & $0.838$ & $0.019$ & $0.014$\\\hline
$500$ & $0.844$ & $0.017$ & $0.015$\\\hline
\end{tabular}
\label{t:cox_sample_size}%
\end{table}%

\subsection{Experiment 3}

\begin{figure}
[ptb]
\begin{center}
\includegraphics[
height=3.8631in,
width=2.9992in
]%
{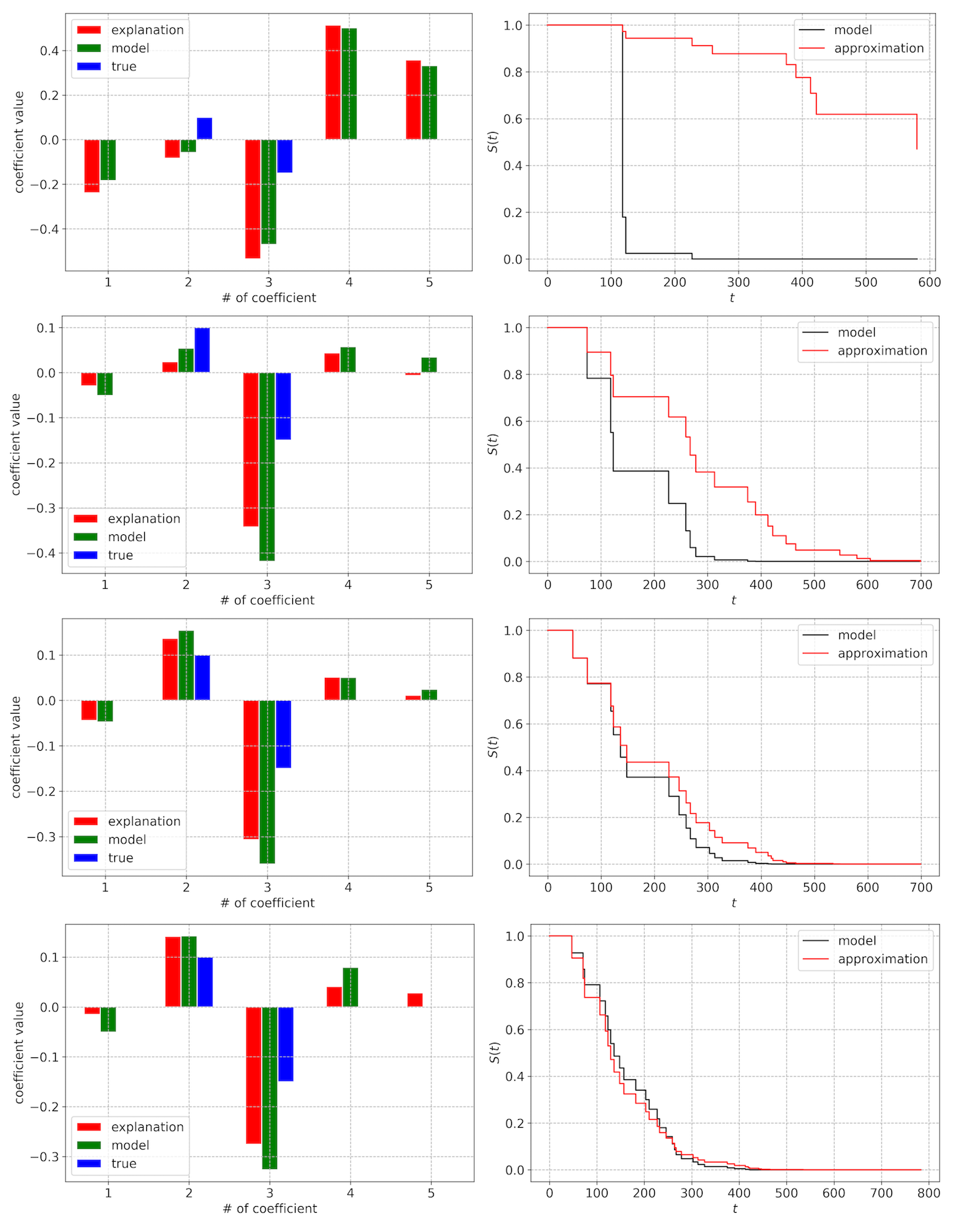}%
\caption{Comparison of important features (left pictures) and survival
functions (right pictures) for the black-box Cox model by $10$, $20$, $30$,
$40$ training examples}%
\label{fig:cox_10}%
\end{center}
\end{figure}

Another interesting question for studying is how SurvLIME behaves by a very
small amount of training data. We use the Cox model and the RSF as black-box
models and train them on $10$, $20$, $30$, $40$ examples from cluster 0. The
models are tested on $10$ examples from cluster 0. Results of experiments for
the Cox model are shown in Fig. \ref{fig:cox_10} where rows correspond to
$10$, $20$, $30$, $40$ training examples, respectively, left pictures
illustrate relationships between important features, right pictures show
survival functions obtained from the black-box Cox model and from the Cox
approximation (see similar pictures in Figs. \ref{cox_sample_size}). It is
interesting to points out from Fig. \ref{fig:cox_10} that important features
obtained by means of SurvLIME differ from the parameters of the Cox model for
generating random examples in the case of $10$ training examples. However,
they almost coincide with features obtained by the black-box model.
Moreover, it follows from Fig. \ref{fig:cox_10} that important features of
SurvLIME quickly converge to parameters of the Cox model for generating random examples.%

The C-index, measures $RMSE_{\text{model}}$ and $RMSE_{\text{true}}$ as
functions of the sample size are provided in Table \ref{t:cox_10}. It can be
seen from Table \ref{t:cox_10} that $RMSE_{\text{model}}$ and
$RMSE_{\text{true}}$ are strongly reduced with increase of the sample size.%

\begin{table}[tbp] \centering
\caption{Approximation accuracy measures for four cases of using the black-box Cox model by the small amount of data}%
\begin{tabular}
[c]{cccc}\hline
& C-index & $RMSE_{\text{model}}$ & $RMSE_{\text{true}}$\\\hline
$10$ & $0.444$ & $0.231$ & $0.293$\\\hline
$20$ & $0.489$ & $0.194$ & $0.197$\\\hline
$30$ & $0.622$ & $0.054$ & $0.082$\\\hline
$40$ & $0.733$ & $0.047$ & $0.053$\\\hline
\end{tabular}
\label{t:cox_10}%
\end{table}%

\begin{figure}
[ptb]
\begin{center}
\includegraphics[
height=3.5743in,
width=2.7492in
]%
{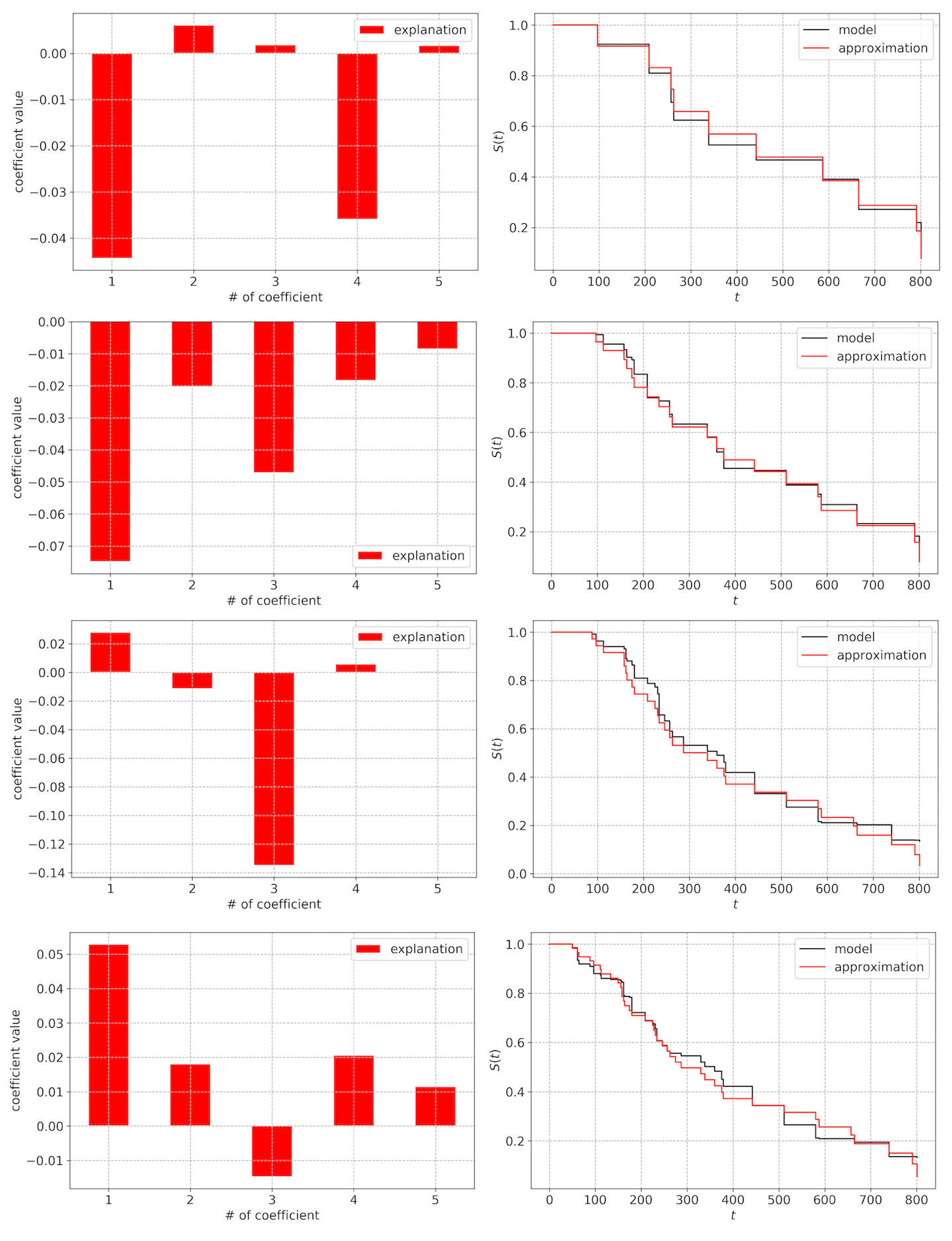}%
\caption{Important features (left pictures) and survival functions (right
pictures) for the black-box RSF by $10$, $20$, $30$, $40$ training examples}%
\label{fig:rsf_10}%
\end{center}
\end{figure}

Results of experiments for the RSF are shown in Fig. \ref{fig:rsf_10} where
rows correspond to $10$, $20$, $30$, $40$ training examples, respectively,
left pictures illustrate important features, right pictures show survival
functions obtained from the black-box RSF and from the Cox approximation
(see similar pictures in Fig. \ref{fig:cox_10}). It can be seen from Fig.
\ref{fig:rsf_10} that the difference between survival functions is reduced
with increase the training set. However, in contrast to the black-box Cox
model (see Fig. \ref{fig:cox_10}), important features are very unstable. They
are different for every training sets. A reason of this important feature
behavior is that they explain the RSF outputs which are significantly changed
by the so small numbers of training examples.%

\section{Numerical experiments with real data}

In order to illustrate SurvLIME, we test it on several well-known real
datasets. A short introduction of the benchmark datasets are given in Table
\ref{t:SurvLIME_datasets} that shows sources of the datasets (R packages), the
number of features $d$ for the corresponding dataset, the number of training
instances $m$ and the number of extended features $d^{\ast}$ using hot-coding
for categorical data.%

\begin{table}[tbp] \centering
\caption{A brief introduction about datasets}%
\begin{tabular}
[c]{cccccc}\hline
Data set & Abbreviation & R Package & $d$ & $d^{\ast}$ & $m$\\\hline
Chronic Myelogenous Leukemia Survival & CML & multcomp & $5$ & $9$ &
$507$\\\hline
NCCTG Lung Cancer & LUNG & survival & $8$ & $11$ & $228$\\\hline
Primary Biliary Cirrhosis & PBC & survival & $17$ & $22$ & $418$\\\hline
Stanford Heart Transplant & Stanford2 & survival & $2$ & $2$ & $185$\\\hline
Trial Of Usrodeoxycholic Acid & UDCA & survival & $4$ & $4$ & $170$\\\hline
Veterans' Administration Lung Cancer Study & Veteran & survival & $6$ & $9$ &
$137$\\\hline
\end{tabular}
\label{t:SurvLIME_datasets}%
\end{table}%

\begin{figure}
[ptb]
\begin{center}
\includegraphics[
height=3.3382in,
width=3.2413in
]%
{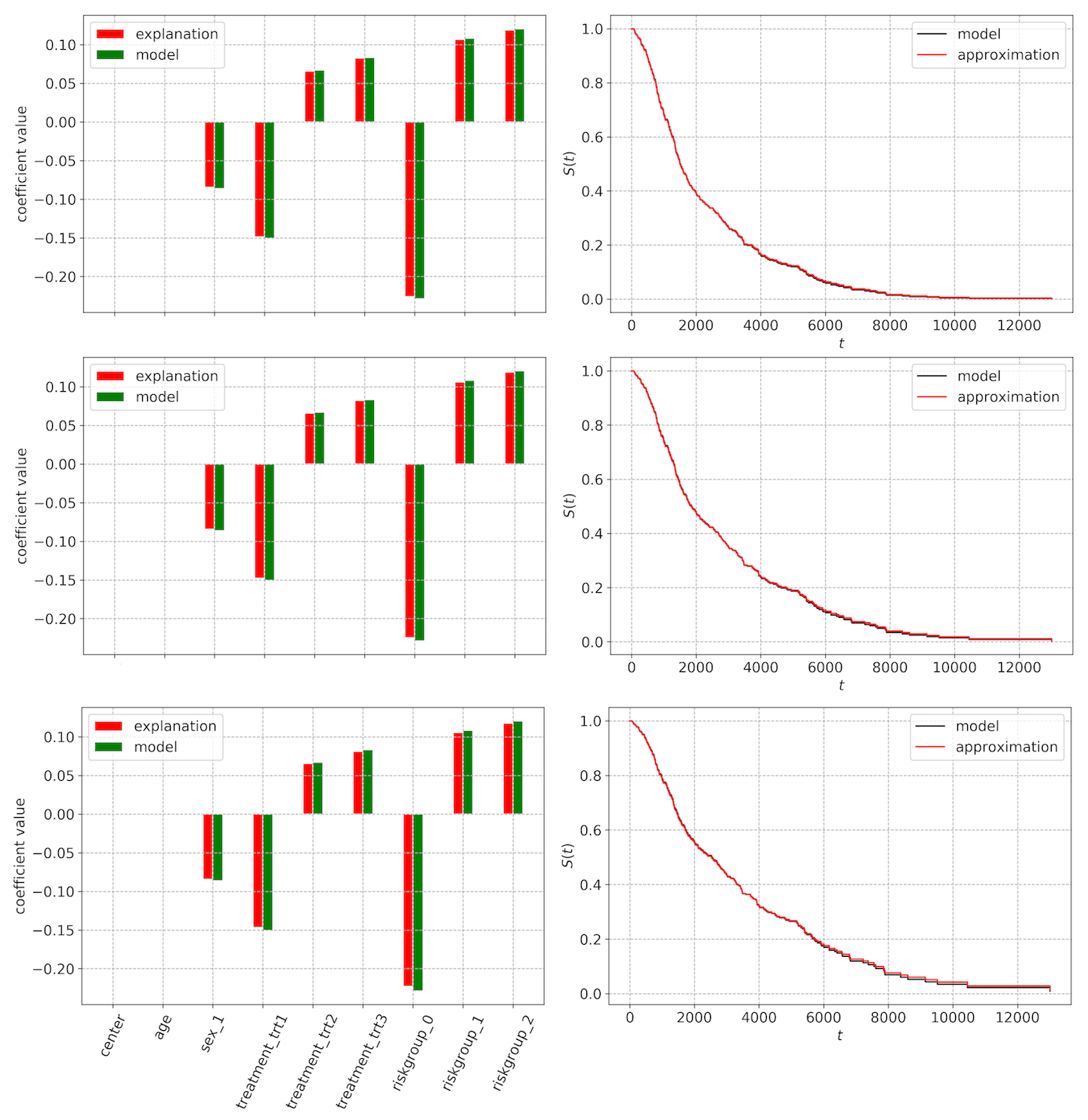}%
\caption{The best, mean and worst approximations for the Cox model trained on
the CML dataset}%
\label{fig:cml_cox}%
\end{center}
\end{figure}

\begin{figure}
[ptb]
\begin{center}
\includegraphics[
height=3.0511in,
width=3.2093in
]%
{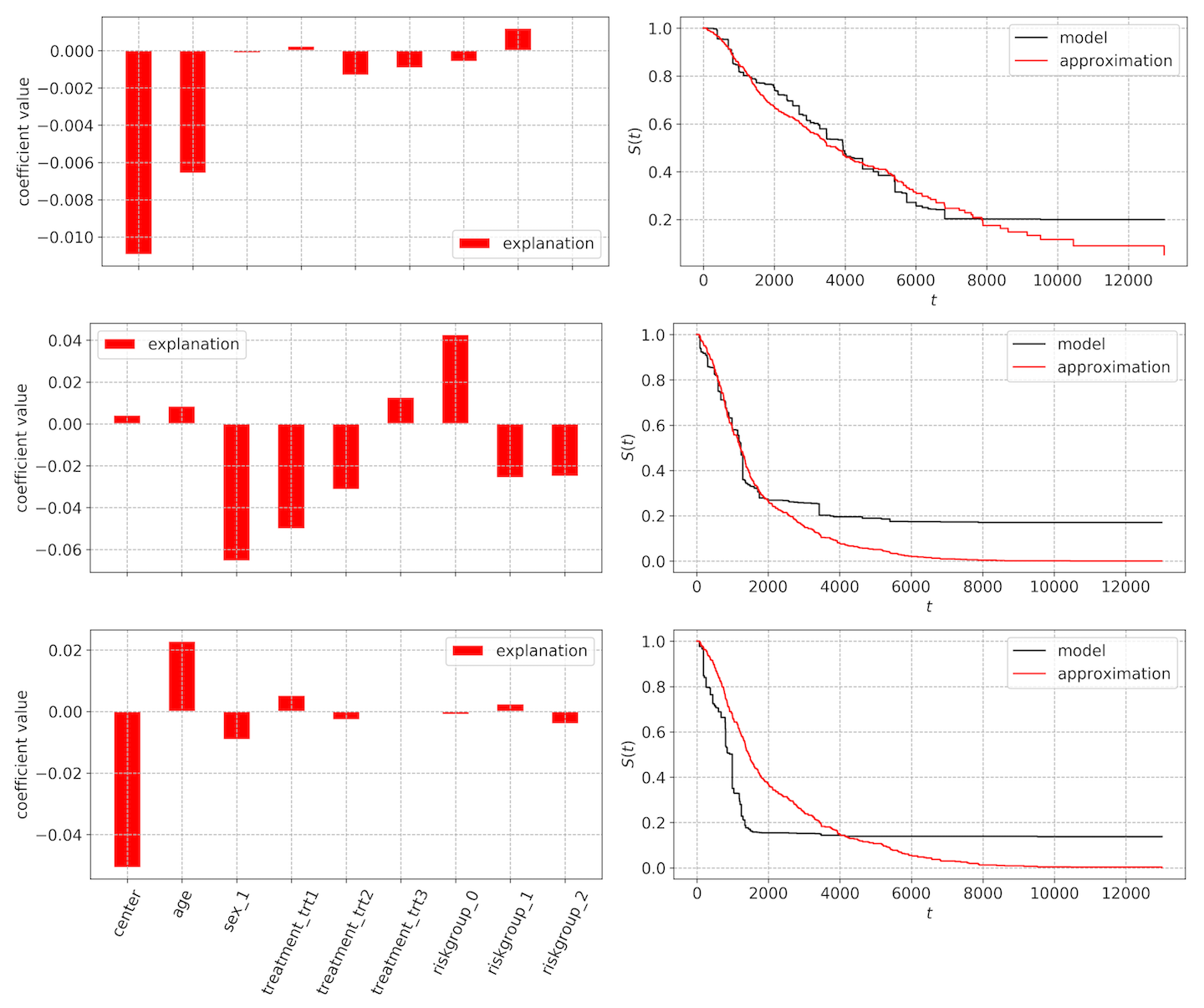}%
\caption{The best, mean and worst approximations for the RSF trained on the
CML dataset}%
\label{fig:cml_rsf}%
\end{center}
\end{figure}

Figs. \ref{fig:cml_cox}-\ref{fig:cml_rsf} illustrate numerical results for the
CML dataset. Three cases of approximation are considered: best (pictures in
the first row), mean (pictures in the second row) and worst (pictures in the
third row). These cases are similar to cases studied for synthetic data. The
cases are studied for the black-box Cox model (Fig. \ref{fig:cml_cox}) and
the black-box RSF (Fig. \ref{fig:cml_rsf}). Again left pictures in figures
show values of important features $\mathbf{b}^{\text{model}}$ and
$\mathbf{b}^{\text{true}}$ for the Cox model and $\mathbf{b}^{\text{true}}$
for the RSF, right pictures illustrate the approximated survival function and
the survival function obtained by the explained model.

\begin{figure}
[ptb]
\begin{center}
\includegraphics[
height=1.5636in,
width=5.3644in
]%
{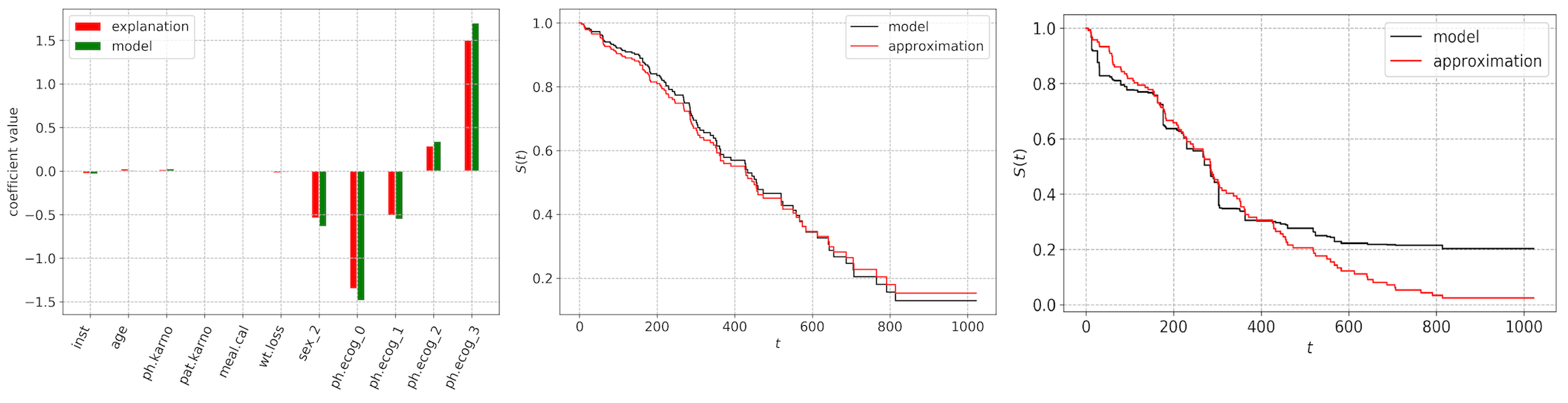}%
\caption{The mean approximation for the Cox model (the first and the second
picture) and the RSF (the third picture) trained on the LUNG dataset}%
\label{f:lung_cox_rsf}%
\end{center}
\end{figure}

\begin{figure}
[ptb]
\begin{center}
\includegraphics[
height=1.5255in,
width=5.3956in
]%
{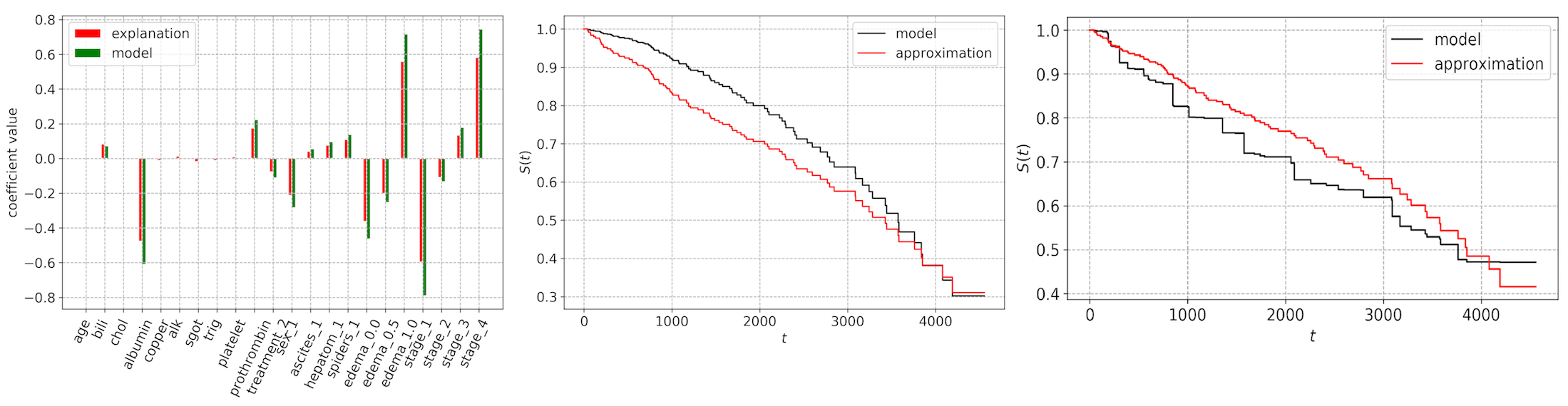}%
\caption{The mean approximation for the Cox model (the first and the second
picture) and the RSF (the third picture) trained on the PBC dataset}%
\label{f:pbc_cox_rsf}%
\end{center}
\end{figure}

\begin{figure}
[ptb]
\begin{center}
\includegraphics[
height=1.5169in,
width=5.495in
]%
{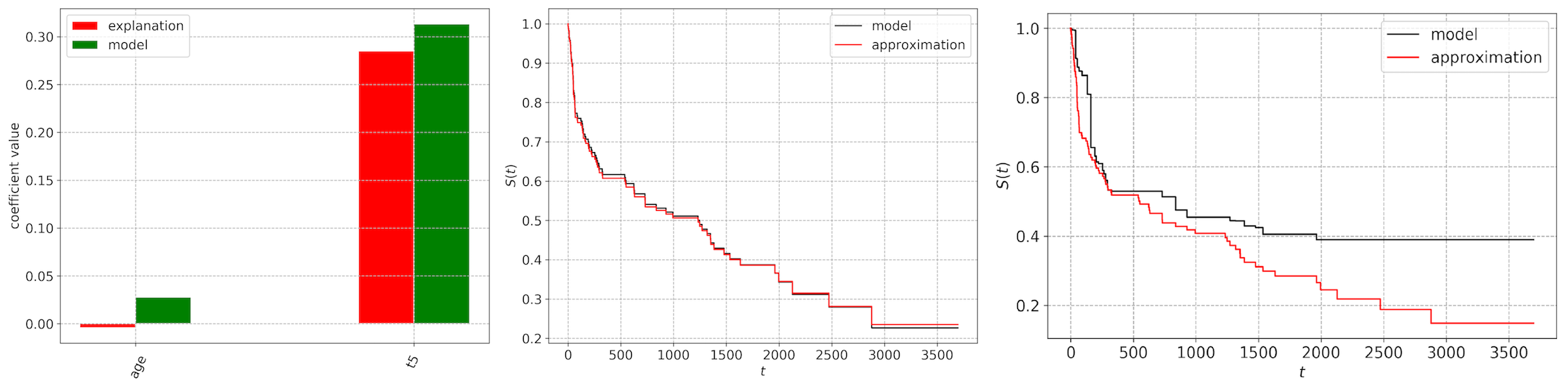}%
\caption{The mean approximation for the Cox model (the first and the second
picture) and the RSF (the third picture) trained on the Stanford2 dataset}%
\label{f:stanford2_cox_rsf}%
\end{center}
\end{figure}

\begin{figure}
[ptb]
\begin{center}
\includegraphics[
height=1.6847in,
width=5.5521in
]%
{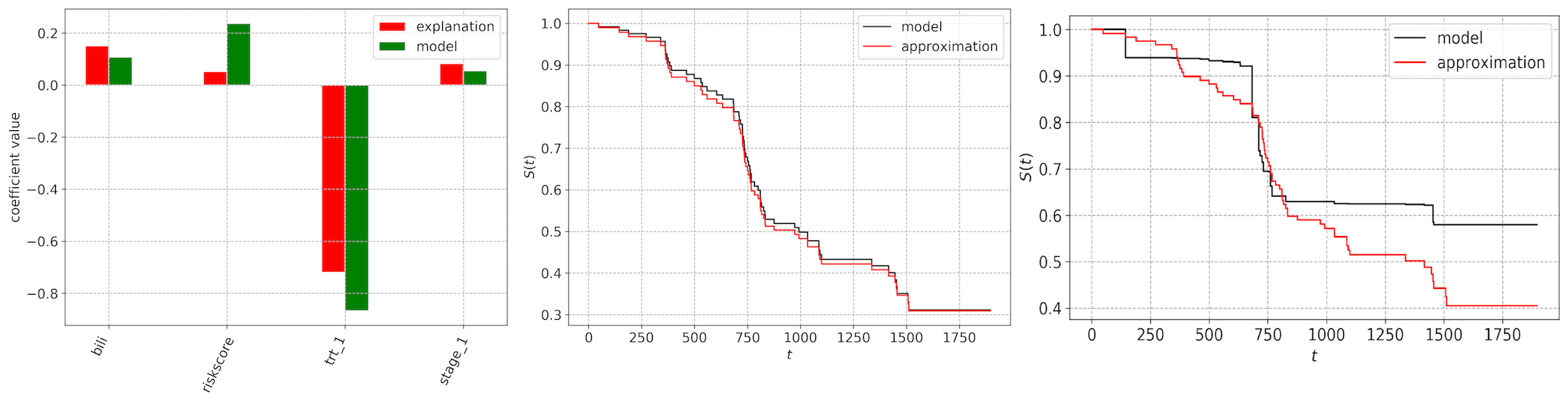}%
\caption{The mean approximation for the Cox model (the first and the second
picture) and the RSF (the third picture) trained on the UDCA dataset}%
\label{f:udca1_cox_rsf}%
\end{center}
\end{figure}

\begin{figure}
[ptb]
\begin{center}
\includegraphics[
height=1.9078in,
width=5.5789in
]%
{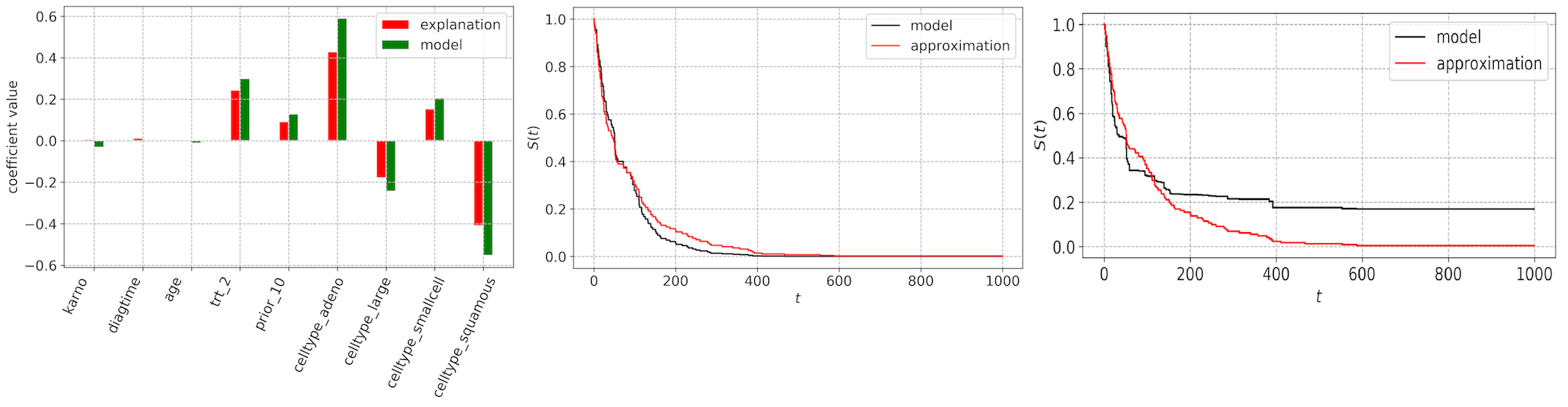}%
\caption{The mean approximation for the Cox model (the first and the second
picture) and the RSF (the third picture) trained on the Veteran dataset}%
\label{f:veteran_cox_rsf}%
\end{center}
\end{figure}

Figs. \ref{f:lung_cox_rsf}-\ref{f:veteran_cox_rsf} show numerical results for
other datasets. Since most results are very similar to the same results
obtained for the CML datasets, then we provide only the case of the mean
approximation for every dataset in order to reduce the number of similar
pictures. Moreover, we do not show important features explaining RSFs because,
in contrast to the Cox model, they cannot be compared with true features.
Every figure consists of three pictures: the first one illustrates the
explanation important features and important features obtained from training
the Cox model; the second picture shows two survival functions for the Cox
model; the third picture shows two survival functions for the RSF.

If the Cox model is used for training, then one can see an explicit
coincidence of explaining important features and the features obtained from
the trained Cox model for all datasets. This again follows from the fact that
SurvLIME does not explain a dataset. It explains the model, in particular, the
Cox model. We also can observe that the approximated survival function is very
close to the survival function obtained by the explained Cox model in all
cases. The same cannot be said about the RSF. For example, one can see from
Fig. \ref{fig:cml_rsf} that the important features obtained by explaining the
RSF mainly do not coincide. The reason is a difference of results provided by
the Cox model and the RSF. At the same time, it can be seen from Figs.
\ref{fig:cml_rsf}-\ref{f:veteran_cox_rsf} that the survival functions obtained
from the RSF and the approximating Cox model are close to each other for many
models. This implies that SurvLIME provides correct results.%

%

%

%

%

%

%

\section{Conclusion}

A new explanation method called SurvLIME which can be regarded as a
modification of the well-known method LIME for survival data has been
presented in the paper. The main idea behind the method is to approximate a
survival machine learning model at a point by the Cox proportional hazards
model which assumes a linear combination of the example covariates. This
assumption allows us to determine the important features explaining the
survival model.

In contrast to LIME and other explanation methods, SurvLIME deals with
survival data. It is not complex from computational point of view. Indeed, we
have derived a simple convex unconstrained optimization problem whose solution
does not meet any difficulties. Moreover, many numerical experiments with
synthetic and real datasets have clearly illustrated accuracy and correctness
of SurvLIME. It has coped even with problems characterizing by small datasets.

The main advantage of the method is that it opens a door for developing many
explanation methods taking into account censored data. In particular, an
interesting problem is to develop a method explaining the survival models with
time-dependent covariates. This is a direction for further research. Only the
quadratic norm has been considered to estimate the distance between two CHFs
and to construct the corresponding optimization problem. However, there are
other distance metrics which are also interesting with respect to constructing
new explanation methods. This is another direction for further research. An
open and very interesting direction is also the counterfactual explanation
with censored data. It could be a perspective extension of SurvLIME. It should
be noted that SurvLIME itself can be further investigated by considering its
different parameters, for example, the assignment of weights of perturbed
examples in different ways, the robustness of the method to outliers, etc. The
original Cox model has several modifications based on the Lasso method. These
modifications could improve the explanation method in some applications, for
example, in medicine applications to take into account a high dimensionality
of survival data. This is also a direction for further research.

\section*{Acknowledgement}

The reported study was funded by RFBR, project number 20-01-00154.


\end{document}